\pdfoutput=1
\documentclass[remotesensing,article,submit,moreauthors]{Definitions/mdpi} 

\usepackage{threeparttable}
\usepackage{array}
\usetikzlibrary{shapes.geometric, arrows.meta, positioning}

\newcolumntype{\$}{>{\global\let\currentrowstyle\relax}}
\newcolumntype{^}{>{\currentrowstyle}}

\newcommand{\bsAdjAddedUnits}{0}

\newcommand{\bsAdjDirect}{3,620}
\newcommand{\bsAdjDirectConfirmed}{737}
\newcommand{\bsAdjDirectHardNegative}{2,883}
\newcommand{\bsAdjDirectPct}{16.1}

\newcommand{\bsAdjPropRule}{majority}
\newcommand{\bsAdjPropagated}{18,756}

\newcommand{\bsAdjTotal}{22,421}

\newcommand{\bsArmLowestBurdenCkpt}{013}

\newcommand{\bsArmLowestBurdenValRecall}{0.434}
\newcommand{\bsArmMaxFOneCkpt}{014}
\newcommand{\bsArmMaxFOneEps}{0.20}

\newcommand{\bsArmMaxFOneGeoPre}{0.522}

\newcommand{\bsArmMaxFOnePostNeg}{0.832}
\newcommand{\bsArmMaxFOnePostPos}{0.867}

\newcommand{\bsArmMaxPrecFlooredTestRecall}{0.563}

\newcommand{\bsArmMaxRecallCkpt}{030}

\newcommand{\bsArmMaxRecallTestRecall}{0.887}
\newcommand{\bsArmMaxRecallThresh}{0.040}

\newcommand{\bsArmMaxRecallValRecall}{0.805}

\newcommand{\bsBurdenSpan}{67}
\newcommand{\bsCollapseCut}{0.05}

\newcommand{\bsCredMaxGeotagsPerCluster}{1.500}
\newcommand{\bsCredMaxMedianExtent}{1.000}
\newcommand{\bsDedupCm}{3}

\newcommand{\bsGmcfName}{GMCF}

\newcommand{\bsMeanDPrec}{-0.176}
\newcommand{\bsMeanDRecall}{+0.231}

\newcommand{\bsPreregArms}{3}
\newcommand{\bsPreregCkpts}{11}
\newcommand{\bsPreregConfigs}{12}

\newcommand{\bsRecallSpan}{3.0}

\newcommand{\bsReviewDedupCm}{3}
\newcommand{\bsReviewEps}{0.20}
\newcommand{\bsReviewMatchRadius}{0.35}
\newcommand{\bsReviewMinSamples}{10}

\newcommand{\bsScreenBlockFrames}{500}
\newcommand{\bsScreenCkpts}{30}
\newcommand{\bsScreenCollapsed}{11}
\newcommand{\bsScreenExcluded}{19}
\newcommand{\bsScreenFlights}{3}
\newcommand{\bsScreenFrames}{1,500}
\newcommand{\bsScreenHealthy}{11}
\newcommand{\bsScreenInView}{348}
\newcommand{\bsScreenSparse}{8}
\newcommand{\bsScreenThresh}{0.03}
\newcommand{\bsSparseCut}{1.7}

\newcommand{\bsSpearman}{+0.007}
\newcommand{\bsSurveyDensityRatio}{3.253}
\newcommand{\bsSweepConfigs}{14,400}
\newcommand{\bsSweepCredible}{3,997}

\newcommand{\bsSweepNoHitCluster}{1,484}

\newcommand{\bsTestDensity}{0.071}
\newcommand{\bsTestFrames}{2,117}
\newcommand{\bsTestGeotags}{161}
\newcommand{\bsTestInView}{151}

\newcommand{\bsValDensity}{0.232}

\newcommand{\bsLamTarget}{100}

\newcommand{\bsSupTraceTen}{108.61}
\newcommand{\bsSupLmaxTen}{74.92}
\newcommand{\bsSupLmaxTwenty}{26.36}

\newcommand{\bsGapTen}{1.450}
\newcommand{\bsGapTwenty}{1.761}

\newcommand{\bsKappaTraceTen}{0.9207}
\newcommand{\bsKappaTraceTwenty}{2.154}
\newcommand{\bsKappaExactTen}{1.335}
\newcommand{\bsKappaInvTen}{1.34}
\newcommand{\bsSigmaZTen}{49.12}
\newcommand{\bsSigmaZTwenty}{98.25}
\newcommand{\bsSigmaYawTen}{0.3013}

\makeatletter
\let\linenumbers\relax
\renewcommand{\contentleftcolumn}{}
\renewcommand{\@maketitle}{%
	\begin{flushleft}
	\vspace*{-1.75cm}
	{\fontsize{18}{20}\selectfont\hyphenpenalty=10000\tolerance=1000\boldmath\bfseries\@Title\par}
	\vspace{12pt}
	{\fontsize{10}{12}\selectfont\boldmath\bfseries\@Author\par}
	\vspace{12pt}
	\end{flushleft}
}
\fancypagestyle{plain}{\fancyhf{}\fancyfoot[C]{\fontsize{9}{9}\selectfont\thepage}}
\fancyheadoffset[L]{0pt}\fancyfootoffset[L]{0pt}
\makeatother

\Title{Human-in-the-Loop Geospatial Annotation for Rapid Dataset Construction in Field-Deployed UAV Systems}

\Author{Morgan Masters $^{1,}$*\orcidA{}, Adam Korycki$^{1}$, Nikolaas Bender$^{2}$, T. Luca Altaffer$^{3}$, Nick Kuipers$^{1}$, Colleen Josephson$^{1}$, and Steve McGuire $^{1}$}

\AuthorNames{Morgan Masters, Nikolaas Bender, T. Luca Altaffer, Nick Kuipers, Colleen Josephson, and Steve McGuire}

\address{%
$^{1}$ \quad Dept. Electrical and Computer Engineering, University of California Santa Cruz\\
$^{2}$ \quad Shield AI; Nikolaas.bender@shield.ai\\
$^{3}$ \quad Pronto AI; luca.altaffer@pronto.ai}

\corres{Correspondence: mwmaster@ucsc.edu}

\thirdnote{All authors were affiliated with the University of Santa Cruz when contributing to this project.}

\abstract{
Real-world perception systems must adapt to changing environments, but manual image annotation cannot scale to field data volumes. 
We present BirdsEye, which shifts expert annotation from images to the field: an operator records target locations in world coordinates using RTK positioning and calibrated projective geometry propagates each observation to all frames where the target is visible. 
To quantify how well physical annotations align with image observations, we derive a first-order mapping from camera-pose uncertainty to pixel uncertainty and validate it against Monte Carlo simulation.
Because that mapping is linear in the six per-axis pose variances, it inverts into a sensor design tool: we give a sufficient condition converting an annotation tolerance into a convex set of admissible pose-noise budgets, a closed-form largest admissible scaling of a deployed sensor suite, and a unique per-axis pose specification under an equal-budget-share allocation.
The condition is conservative by at most a factor of two, which we show matters only for platforms sitting near the tolerance boundary; for those, the exact worst-case eigenvalue condition settles the question, and we report both for our own hardware.
We also analyze the planar-surface approximation underlying the projection, which holds up to 10 degrees of terrain slope.
By direct measurement, we show that system projection accuracy is sub-decimeter (sub-30 pixel) at AGL altitudes of 10-20\;m under conditions excluding sustained yawing.
During an in-field case study across three agricultural sites, two field workers produced 12,524 annotated frames carrying 55,600 labels in roughly 12 hours (25.5$\times$ per-worker rate increase over manual labeling).
Detectors trained on imagery collected by this workflow recovered 56–89\% of in-view surveyed targets at a geographically distinct farm, at pre-registered operating points; human review of the leading configuration estimates detection precision at 83–87\%, spanning three tie-break conventions for clusters carrying contradictory human verdicts.
}


\addhighlights{yes}
\renewcommand{\addhighlights}{

\noindent\textbf{What are the main findings?}
\begin{itemize}[labelsep=2.5mm,topsep=-3pt]
\item Recording targets once in world coordinates with RTK positioning and projecting them into UAV imagery let two field workers label 12,524 frames in about 12 hours, a 25.5x per-worker gain over manual image labeling, with sub-decimeter projection error at 10-20m AGL.
\item A first-order model of pose-to-pixel uncertainty, validated against Monte Carlo simulation, is linear in the six per-axis pose variances and inverts into admissible sensor noise budgets, a closed-form suite scaling factor, and a unique per-axis pose specification.
\end{itemize}\vspace{3pt}
\textbf{What are the implications of the main findings?}
\begin{itemize}[labelsep=2.5mm,topsep=-3pt]
\item Domain experts can build fine-resolution aerial training sets for targets missing from existing corpora by surveying in the field instead of in images, and detectors trained this way transferred to an unseen farm under pre-registered evaluation.
\item Platform designers can size pose sensors against an annotation tolerance before building hardware, screening with the linear bound and confirming near-boundary designs with the exact eigenvalue condition, within a planar-terrain envelope of roughly 10 degrees slope.
\end{itemize}
}

\begin{document}


\section{Introduction}
\label{sec:intro}
 
Drone-borne deep computer vision models are increasingly used for persistent environmental monitoring, particularly in remote or inaccessible places \cite{bhardwaj2016uavs, Maes2019Feb, klemas2015coastal, neupane_automatic_2021, pettorelli_using_2005, adao_hyperspectral_2017, honrado_uav_2017, moriya_detection_2021}.
Problematically, the performance of many deep-learned perception methods hinges on the availability of large-scale datasets to learn patterns that humans recognize intuitively.
The annotation stage of dataset assembly is famously time-consuming and costly~\cite{Yalalov2023Feb}, forcing research communities to rely on large-scale, internet-hosted datasets for robust AI performance.
 
As a consequence, researchers and practitioners working in specialized domains---such as field robotics for environmental monitoring or agricultural automation---often find that their specific data needs are underrepresented~\cite{Kamilaris2018AgriDL}.
For instance, numerous well-annotated plant health datasets exist, with sizes ranging from several hundred to tens of thousands of labeled examples \cite{Mohanty2016PlantVillage, Xie2022CropDisease, Xu2024Sep}, but these are predominantly captured from close-up perspectives.
A farmer attempting to manage a 500-acre (2 km$^2$) operation requires data views from more practical perspectives, like that of an uncrewed aerial vehicle (UAV), for effective field assessment.
The difficulty is that the symptoms such an operator is tracking---lesions, emergence gaps, burrow entrances---are often only millimeters to centimeters across.
Resolving them from the air imposes a ground sample distance (GSD, the real-world extent of a pixel) requirement that in turn constrains the operating altitude, and it is precisely this low-altitude, fine-resolution viewpoint that
the available corpus does not cover.
 
Large aerial datasets do exist \cite{Zhu2018Apr, xia2018dota, SemanticDroneDataset2019}, but their contents reflect the priorities that funded them: structured urban scenes, vehicles, and pedestrians, imaged at altitudes and ground resolutions suited to those subjects.
The same imbalance appears in the georeferencing literature. 
Reported accuracies for direct georeferencing of UAV imagery are in the range of 8-11cm~\cite{liu2022accuracy, turner2013direct}, obtained from imagery with GSDs of similar scale.
Those figures are entirely adequate for the targets those studies address, and insufficient for a target a few millimeters across. 
The gap is therefore not only one of subject matter but of spatial fidelity.

 \begin{figure}[t]
    \centering
    \includegraphics[width=\linewidth]{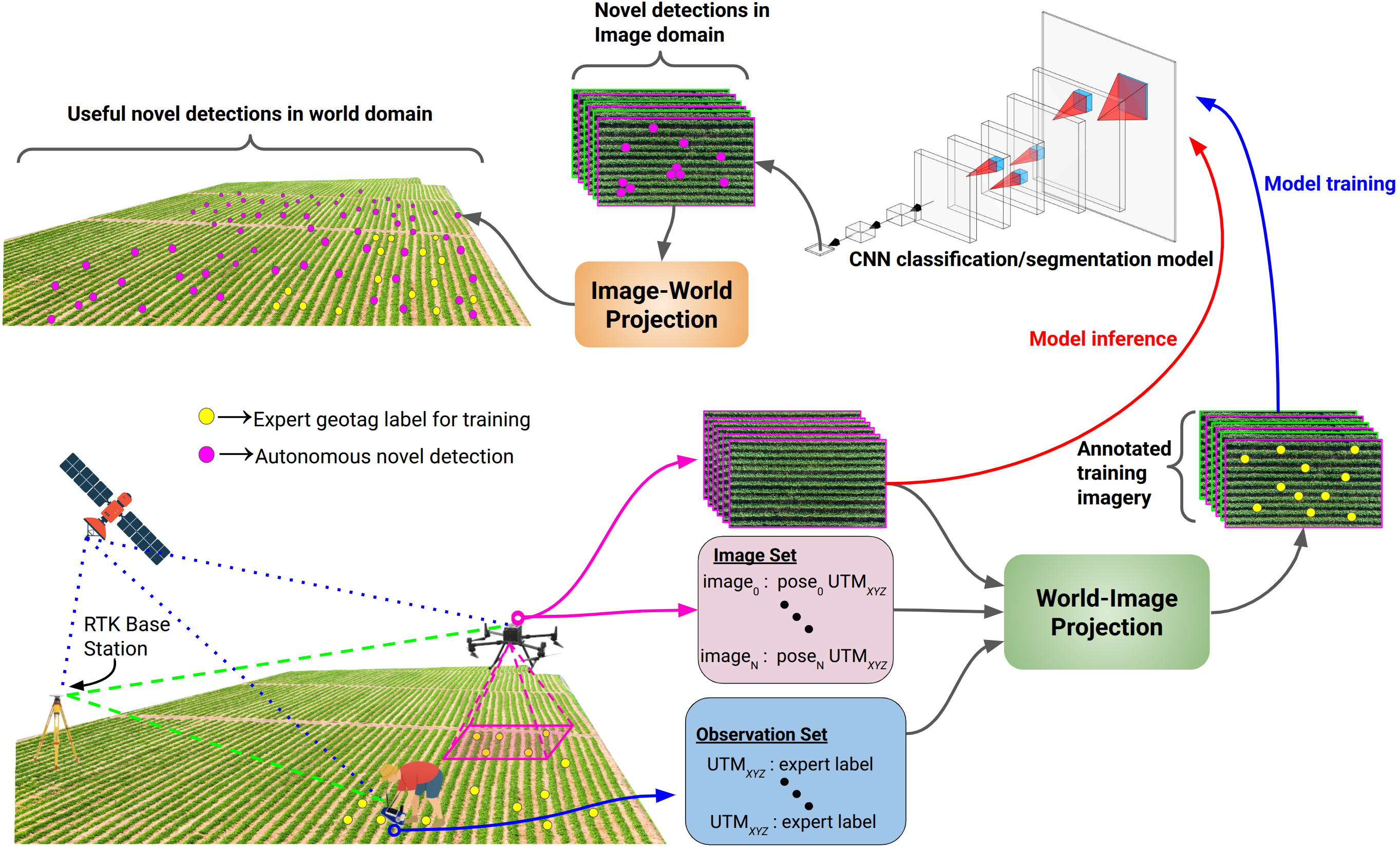}
    \caption{The \textit{BirdsEye} UAV-based image annotation system. Pictured (bottom left) are the specialist in the field with a handheld annotation device, the RTK GPS base station, and the UAV surveying a field. System software (bottom right) then either collates expert annotations and UAV imagery via camera projective models, creating an annotated dataset, or passes UAV imagery directly to a trained neural model for feature detection and subsequent georeferencing of detection event pixels (top left).}
    \label{fig:birdseye_flow}
\end{figure}  

In response, we introduce BirdsEye, a UAV-based framework that moves expert annotation from image coordinates into the physical environment (Figure~\ref{fig:birdseye_flow}). 
Rather than requiring an annotator to inspect and label individual aerial frames, a field operator records the locations of features of interest directly in the environment using RTK positioning. 
These spatial annotations are subsequently transformed into the coordinate frame of the UAV camera and projected into the acquired image stream, automatically producing image-space labels for supervised learning. 
The same geometric representation permits image-space detections to be mapped back into world coordinates, enabling a common spatial representation for both dataset construction and subsequent deployment.
 
This coordinate-system inversion changes the annotation workflow from image-centric labeling to field-centric observation. 
The distinction is important for field robotics, where domain experts may be able to identify relevant phenomena efficiently while physically present in the environment, but may not be efficient image annotators when presented with thousands of aerial frames. 
BirdsEye therefore separates the act of identifying a target from the act of associating that target with individual sensor observations. 
The former is performed in the physical environment by a domain expert; the latter is
automated through calibrated projective geometry.
 
Our method raises a geometric question: how accurately can a physical annotation be associated with an image observation given uncertainty in the platform pose and imperfections in the assumed scene geometry? We therefore formulate the projection process explicitly and derive a first-order propagation of camera pose uncertainty into pixel-space uncertainty.
The resulting analytical model provides an estimate of annotation uncertainty directly from the pose covariance and camera geometry, and inverts to give the pose precision a given annotation tolerance demands. 
We evaluate this approximation against direct Monte Carlo simulation and experimentally measured projection errors under controlled flight maneuvers.

Because that propagation is linear in the individual pose variances, it also runs in the opposite direction: from a required annotation tolerance to the sensor noise a platform must achieve to meet it. We develop this inverse as an explicit design condition rather than a single scalar figure of merit. 
The admissible noise budgets form a convex polyhedron whose bounding hyperplane is set by the imaging geometry at the worst-imaged point in the frame, so screening a candidate sensor suite is one linear evaluation rather than a search over pixels, and the largest uniform inflation of an existing suite follows in closed form. 
Selecting a single specification out of that set requires a stated preference; we adopt the allocation in which every degree of freedom contributes the same worst-case pixel variance, which is unique and scale-invariant and therefore states a requirement on a platform rather than describing the one we fly. 
The condition is sufficient, not necessary, and conservative by at most a factor of two; we identify the regime in which that slack changes an engineering decision and where the exact worst-case condition should be used to confirm a near-boundary platform.
 
Finally, we evaluate the complete system in field deployments across multiple agricultural sites, in a case study on the detection of burrowing pest activity.
The total campaign dataset comprises 46,205 aerial frames carrying 122,467 field annotations, across two deployment sets, with one geographically distinct site held out for evaluating trained detection models (Table~\ref{tab:valflight-summary} Upper, Partner Site 1; 2,120 frames, 6,354 annotations). 
Because a detector reported on its own training distribution establishes very little, the operating points at which that model is scored were fixed and sealed before the held-out site was processed, and every resulting detection was subsequently reviewed by a human. 
This deployment demonstrates the practical consequence of the proposed coordinate-system inversion: substantial reductions in human image-labeling effort while retaining a direct connection between expert observations, image-space training labels, and world-referenced detections.
 
Our key contributions are therefore:
\begin{itemize}
\item A human-in-the-loop geospatial annotation framework that represents human
observations directly in world coordinates and automatically transforms them
into image-space training labels for UAV imagery.
 
\item A multi-site field deployment demonstrating rapid construction of a large
supervised dataset for a target class absent from existing aerial corpora.
 
\item A first-principles analytical model for propagation of camera pose
uncertainty into pixel-space annotation uncertainty, validated against Monte
Carlo simulation and against measured projection error across multiple
altitudes and controlled flight maneuvers.

\item An inversion of that model into a sensor noise design condition: a bound
linear in the six per-axis pose variances that converts a target annotation
tolerance into a convex set of admissible pose-noise budgets, a closed-form
admissible scaling of a deployed suite, and a unique per-axis pose
specification, together with a characterization of the bound's factor-of-two
conservatism and the operating regime in which it is decision-relevant.
 
\item A systematic characterization of the flat-world approximation,
establishing an empirical terrain-slope operating envelope for the current
projection model.
 
\item A pre-registered held-out evaluation, in which operating points were fixed
before the unseen test site was scored and every resulting detection was
subsequently adjudicated by a human reviewer, testing whether the constructed
dataset transfers to an environment that was never used to build or tune
anything.
\end{itemize}

\section{Background and Related Work}
\label{sec:background}

The construction of supervised datasets for field-deployed perception systems is typically formulated as an image-space annotation problem: targets are identified and localized within acquired images, and the resulting image-space labels are used for model training. General-purpose annotation platforms such as Labelbox, CVAT, and Roboflow provide scalable interfaces for this workflow, including dataset management, collaborative labeling, and automated or AI-assisted annotation capabilities \cite{Labelbox2023,CVAT2022,Roboflow2023}. These systems are effective when the image is the natural domain in which an annotator can identify the target. For field robotics and specialized remote-sensing applications, however, domain experts may have substantially better access to the target in the physical environment than in the resulting collection of aerial image frames.

A complementary line of work attempts to reduce the amount of manual image annotation through learned or weakly supervised methods. Weakly supervised semantic annotation has been used to derive object information from incomplete supervision in high-resolution satellite imagery \cite{7414501}, while unsupervised domain adaptation can reduce dependence on densely labeled target-domain imagery \cite{9468936}. Other approaches generate labels automatically using physical markers \cite{8641376} or explicitly study the effects of noisy localization during detector training \cite{9913838}. These methods reduce the amount or precision of required image-space supervision, but the underlying annotation workflow remains centered on observations made in image space.

\subsection{Geospatial and Image-Based Annotation}

Geospatial imagery platforms extend conventional annotation by associating image observations with geographic coordinates. Picterra provides a platform for machine-learning-based analysis of geospatial imagery \cite{Picterra2021}, while LabelMe provides a general web-based framework for image annotation \cite{LabelMe2008}. These systems demonstrate the utility of spatially aware annotation and georeferenced imagery for remote-sensing applications. Related work has also used the geographic distribution of images to guide annotation effort. Yamada et al. \cite{9669070} exploit georeferenced imagery to select representative observations and reduce the amount of imagery requiring human labeling.

The distinction between these approaches and BirdsEye is therefore not the use of geographic coordinates itself. Rather, it is the domain in which the annotation event occurs. Existing georeferenced-image approaches generally retain the image as the fundamental annotation unit: geographic information is used to organize, select, augment, or interpret image observations. BirdsEye reverses this relationship. The fundamental annotation event is an observation made in the physical environment and represented directly in world coordinates; calibrated projective geometry then determines which image observations contain that physical target and generates the corresponding image-space labels. Thus, the system separates the semantic task of identifying a target from the geometric task of associating that target with individual sensor observations.

This distinction is particularly relevant to UAV surveys, in which a single physical target may appear in many consecutive frames. Conventional image-space annotation requires the correspondence between the target and each image to be established repeatedly by a human annotator. In BirdsEye, the physical observation is recorded once and subsequently propagated to every image in which the target is geometrically visible. The resulting image labels remain compatible with conventional supervised-learning pipelines, while the source annotation retains an explicit world-coordinate representation that can also be used to georeference subsequent model detections.

\subsection{Synthetic Datasets}

Synthetic data generation provides another strategy for reducing dependence on manually collected and annotated imagery. Recent work has explored synthetic multimodal datasets specifically for UAV object detection \cite{Yao2024Jun}, while simulation and rendering environments such as Unity Perception \cite{unity-perception2022}, CARLA \cite{Dosovitskiy17}, Gazebo \cite{Koenig}, and Stonefish \cite{stonefish} provide mechanisms for generating synthetic observations with known scene geometry and object labels. These systems can produce large quantities of precisely labeled data without requiring human annotation of every image. Photorealistic rendering and image-domain translation can further reduce the discrepancy between synthetic and observed imagery \cite{richterGTAV}.

The principal limitation of this approach for field-deployed remote sensing is the transfer from simulated to observed imagery. Even highly detailed rendering pipelines must approximate the appearance, illumination, materials, sensor characteristics, and environmental variability encountered in real deployments. Domain gaps between synthetic and real imagery can therefore limit direct transfer of models trained exclusively on simulated data. BirdsEye instead retains real-world image acquisition and derives supervision from observations made directly in the physical environment. It therefore does not eliminate the cost of data collection, but avoids requiring a separate synthetic representation of the target domain and provides labels tied directly to the observed field conditions.

\subsection{Physical and Robotic Label Generation}

A closer conceptual precedent is provided by robotic systems that generate training labels from physical interactions with the environment. Kiyokawa et al. \cite{8641376}, for example, use visual markers attached to physical targets to enable automated annotation during data collection. Their approach demonstrates the value of establishing correspondence between a known physical target and the images acquired by a robot, but requires a physical marker to be placed on each target and subsequently masked during training.

De Gregorio et al. \cite{8844069} similarly investigate semiautomatic labeling for robotic deep-learning applications using interaction with the physical workspace. Their system demonstrates how a human can identify or specify training targets in a physical environment while automation establishes the corresponding image labels. This work is particularly relevant to BirdsEye because it separates target specification from conventional manual image annotation. However, its operating environment is an indoor robotic manipulation setting with a comparatively constrained camera and workspace.

BirdsEye extends this physical-world labeling principle to outdoor UAV remote sensing by representing target observations in a global spatial coordinate system. Rather than requiring physical markers or an indoor augmented-reality workspace, the target is identified through an RTK-referenced field observation. The UAV's measured pose and calibrated camera model then provide the transformation from the world coordinate of the observation to its image-space location. This enables the same physical annotation to be associated with observations from a moving aerial platform across an extended outdoor survey.
The distinction is consequential because the geometric association can be performed automatically for every sensor observation for which the target is visible. It also makes the uncertainty of the correspondence an explicit geometric quantity rather than an implicit property of manual image annotation.
The following sections therefore treat the world-to-image transformation as the central mechanism by which \textit{BirdsEye} changes the annotation workflow.

\subsection{Pixel Georeferencing}
Georeferencing is the task of assigning geospatial coordinates to pixels in aerial or satellite imagery. 
Accurate georeferencing is critical for applications in environmental monitoring and autonomous navigation, and particularly for ultra-fine feature georeferencing tasks (features on the sub-cm scale). 
Two dominant paradigms exist: Ground Control Point (GCP)-based methods and direct georeferencing.

\subsubsection{GCP-Based Georeferencing}
The traditional approach to georeferencing uses GCPs, or visually identifiable landmarks with high-accuracy ground-truth coordinates. 
GCPs are distributed in the survey area and, by establishing correspondences between image pixels and their known ground coordinates, a transformation model (e.g. projective or homography-based) is estimated to map the image plane to a geospatial reference system. 
Liu et al.~\cite{liu2022accuracy} reports an accuracy of 3.5 cm when using 10-12 GCPs per km$^2$. 
However, the GSD in their study was 1.7 cm making this infeasible for ultra-fine grain feature tracking. 

\subsubsection{Direct Georeferencing}
Direct georeferencing offers a contactless alternative to traditional ground control point (GCP)-based methods by eliminating the need for surveyed visual markers. 
Instead, it relies on onboard localization sensors to estimate the camera's six degree-of-freedom pose at the precise moment of image capture. 
This typically involves tightly-coupled GNSS/IMU integration to provide accurate position and orientation estimates. 

Liu et al.~\cite{liu2022accuracy} report an absolute georeferencing accuracy of 8.7 cm using direct methods, while~\cite{turner2013direct} finds a limiting performance of 11 cm $\pm$ 2.14 cm. 
Both studies employ imagery with relatively coarse GSDs, on the order of several centimeters, which limits their ability to resolve small-scale surface features. 
Direct georeferencing applied to imagery with sub-3 mm GSD requires significantly higher precision to avoid pixel-level misalignment. 

\section{Materials and Methods}
\label{sec:methods}

\begin{figure}
    \centering
    \scalebox{0.75}{ 
    \begin{tikzpicture}[node distance=1cm and 2cm]
    
    \tikzstyle{block} = [rectangle, rounded corners, minimum width=3.2cm, minimum height=1cm, text centered, draw=black, fill=blue!20]
    \tikzstyle{source} = [rectangle, rounded corners, minimum width=3.5cm, minimum height=1cm, text centered, draw=black, fill=red!30]
    \tikzstyle{sensor} = [rectangle, rounded corners, minimum width=3.5cm, minimum height=1cm, text centered, draw=black, fill=green!30]
    \tikzstyle{network} = [rectangle, rounded corners, minimum width=3.5cm, minimum height=1cm, text centered, draw=black, fill=orange!30]
    
    \tikzstyle{arrow} = [thick,->,>=stealth]
    \tikzstyle{power} = [thick,dashed,->,>=stealth,color=orange]
    \tikzstyle{data} = [thick,->,>=stealth,color=green!60!black]
    \tikzstyle{trigger} = [thick,->,>=stealth,dotted,color=purple]
    
    \node (drone) [source] {DJI M300 RTK};
    \node (powerboard) [block, below=of drone] {Power Distribution Board};
    \node (rpi) [block, below=of powerboard] {Raspberry Pi 5};
    
    \node (camera) [sensor, right=1.8cm of powerboard, yshift=0.9cm] {GigE RGB Camera};
    \node (ins) [sensor, right=2.2cm of rpi, yshift=1.1cm] {INS};
    
    \node (radar) [sensor, right=of drone, yshift=0.7cm] {Radar Altimeter};
    \node (modem) [network, right=2.2cm of rpi, yshift=-1.8cm] {RTK Radio Modem};
        
    \draw [arrow] (drone) -- (powerboard) node[midway,left] {\footnotesize 24V (Nom.)};
    \coordinate (powermerge) at ([xshift=1cm]powerboard.east);
    \draw [power] (powerboard.east) -- (powermerge);
    \draw [power] (powerboard) -- (rpi) node[midway,left] {\footnotesize 5V};
    \draw [power] ([yshift=0.2cm]powerboard.east) --++ (0.75,0) |- (camera.west) node[midway,above left] {\footnotesize 12V};
    \draw [power] (powermerge) --++ (0,0) |- (ins.west) node[midway,below left] {\footnotesize 5V};
    \draw [power] (powermerge) --++ (0,0) |- (radar.west) node[midway,above right] {\footnotesize 5V};
    \draw [power] (modem.west) -| ([xshift=0.25cm]rpi.south) node[pos=0.3,above] {\footnotesize USB Power};

    \coordinate (datamerge) at ([xshift=1cm]rpi.east);
    \draw [data] (radar.east) --++ (1.5cm,0) --++ (0,-5.7cm) -| (datamerge) node[pos=0.3,above] {\footnotesize USB Serial};
    \draw [data] (camera.east) --++ (1.25cm,0) --++ (0,-3.4cm) -| (datamerge) node[pos=0.3,above] {\footnotesize Ethernet};
    \draw [data] (ins.east) --++ (1cm,0) --++ (0,-1.1cm) -| (datamerge) node[pos=0.3,above] {\footnotesize TTL serial};
    \draw [data] (datamerge) -- (rpi.east);
    \draw [data] ([yshift=-0.05cm]modem.west) -| ([xshift=0.2cm]rpi.south) node[pos=0.3,below] {\footnotesize + Data};
    
    \draw [trigger] (camera.south) -- ([xshift=-0.04cm]ins.north) node[pos=0.5,right] {\footnotesize Strobe Trigger};
    
    \end{tikzpicture}
}
    \vspace{0.25cm}
    \begin{center}
\scalebox{0.75}{
\begin{tikzpicture}[
  node distance=0.5cm and 2cm,
  every node/.style={font=\sffamily},
  process/.style={rectangle, draw=black, rounded corners, minimum width=3cm, minimum height=1cm, text centered, fill=blue!20},
  data/.style={rectangle, draw=black, rounded corners, minimum width=3cm, minimum height=1cm, text centered, fill=green!20},
  logic/.style={rectangle, draw=black, rounded corners, minimum width=3cm, minimum height=1cm, text centered, fill=orange!30},
  output/.style={rectangle, draw=black, rounded corners, minimum width=3cm, minimum height=1cm, text centered, fill=red!30},
  arrow/.style={thick,->,>=Stealth},
  dashedarrow/.style={thick,dashed,->,>=Stealth, color=red}
]

\node[process] (streamer) at (0,0) {Database Streamer};
\node[process, above=of streamer] (parser) {Data Collation};
\node[process, above=of parser] (sync) {Time Sync};

\node[data, right=3cm of sync] (rawdata) {Raw Data};
\node[logic, right=1.2cm of parser] (annotation) {Annotation};
\node[logic, right=1cm of streamer] (detection) {Detection};
\node[output, right=1.2cm of annotation] (training) {Training};
\node[output, right=1.2cm of detection] (mapping) {Mapping};

\draw[arrow] (rawdata.west) -- (sync.east);
\draw[arrow] (sync) -- (parser);
\draw[arrow] (parser) -- (streamer);
\draw[arrow] ([yshift=0.2cm]streamer.east) --++ (0.5cm,0) |- (annotation.west);
\draw[arrow] (streamer.east) --++ (0.5cm,0) |- (detection.west);
\draw[arrow] (annotation.east) -- (training.west);
\draw[arrow] (detection.east) -- (mapping.west);


\end{tikzpicture}
}
\end{center} 
    \caption{(Upper) System diagram of the onboard robotic payload integrated with the DJI M300 RTK. (Lower) The \textit{BirdsEye} software pipeline.}
    \label{fig:system_diagram}
\end{figure}
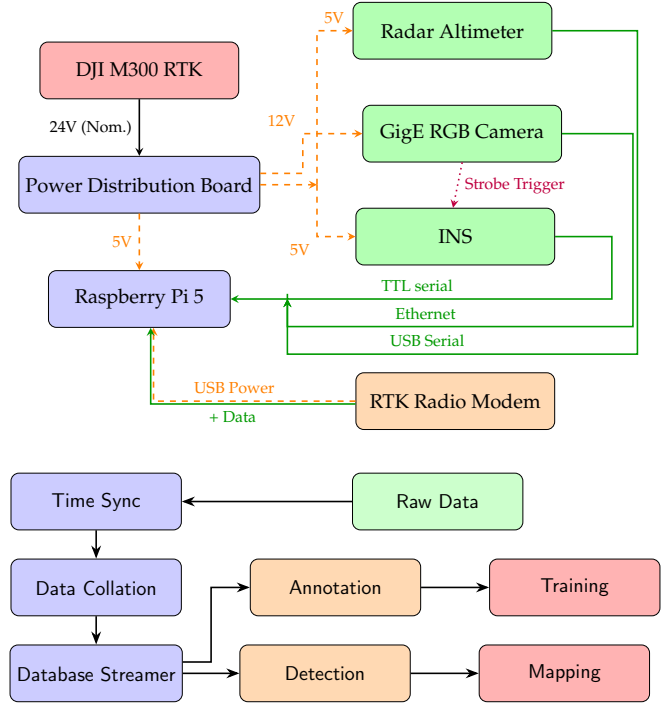

\subsection{The BirdsEye Geoannotation Hardware}
\label{sec:methods_geoannotation}
The \textit{BirdsEye} system employs RTK positioning for centimeter-scale localization precision. 
The geoannotation system we have constructed features an RTK basestation and hand-portable rover units.
Our basestation consists of an L1/L2 GPS receiver, a radio modem, and a Raspberry Pi 4 generating RTCM3 corrections and hosting a Wi-Fi hotspot.
The basestation broadcasts RTCM3 corrections via the radio, for long-range applications, and a local NTRIP server via WiFi to provision a more general set of RTK-compatible wireless devices.
The rover units are handheld annotation devices, featuring the same GPS and radio equipment as the basestation and a user interface.
Every component of the geoannotation system is commercially available, and the NTRIP correction service runs on free and open-source software\footnote{https://github.com/semuconsulting/pygnssutils}.

\subsection{Airframe and Payload Sensors}
\label{sec:methods_payload}
Figure~\ref{fig:system_diagram} illustrates the basic layout of the payload hardware system.
Our airframe is a DJI M300 RTK and the payload computer is a Raspberry Pi 5.
We use a 2MP Gigabit Ethernet (GigE) machine vision camera.\footnote{FLIR Blackfly Machine Vision Camera}
In addition to visual sensing, we include a radar altimeter\footnote{AInstein US-D1 All-Weather Altimeter} and an RTK GPS-provisioned 9-DoF inertial navigation system (INS).\footnote{InertialSense RUG-3-IMX-5-DUAL} 

\subsection{Payload Calibration}
\label{sec:methods_calib}
We used COLMAP~\cite{schonberger2016structure} estimates of Brown-Conrady camera model intrinsic and distortion parameters, computed from the raw imagery of a calibration dataset.
Then, we provided those parameters and the inertial data captured to the Kalibr visual-inertial calibration process~\cite{Furgale2013kalibr} to estimate the rigid body transform between our system's INS and cameras.
These calibration parameters have been set as fixed parameters, and are assumed constant thereafter.

\subsection{Payload Synchronization}
\label{sec:methods_sync}
Our camera is configured to produce strobe pulses at the start of each exposure to trigger 6-DoF pose captures by the INS. 
The resulting sequences of poses and images are easily aligned via the heuristic that the difference in age between the pose and its matching image should never exceed the reciprocal of the framerate. 
If this occurs, we receive a new pose after the capture period and discard the preceding pose-image pair.
Upon finding a match, we remove the paired image from the set of remaining candidates to avoid double-matches.
This combination of hardware support and processing heuristic allows us to reduce latency between imagery and poses, induced by the image transport mechanism, to be on the order of logic delays.

\subsection{Field Operations Protocol}
\label{sec:methods_protocol}

Each deployment starts with powering the RTK base station and the survey handheld devices (Section~\ref{sec:methods_geoannotation}). 
While this occurs, the environment is partitioned into small areas which are easily surveyed by a single person in roughly an hour.
When all survey handhelds show RTK fixes, the field partitions are surveyed sequentially by a team of annotators.
The act of annotation is simply placing the GPS antenna of the survey handheld directly above a found training target instance and triggering a coordinate-and-metadata packet capture.
When a partition's survey completes, records of feature locations are offloaded from the devices to a shortest-path flight planner script, powered by the \verb+fast-tsp+ Python library\footnote{https://fast-tsp.readthedocs.io/en/latest/}, and then the drone is launched on the resulting autonomous flight mission (capturing imagery at 5Hz). 

\begin{figure}[ht]
    \centering
    \includegraphics[width=\linewidth]{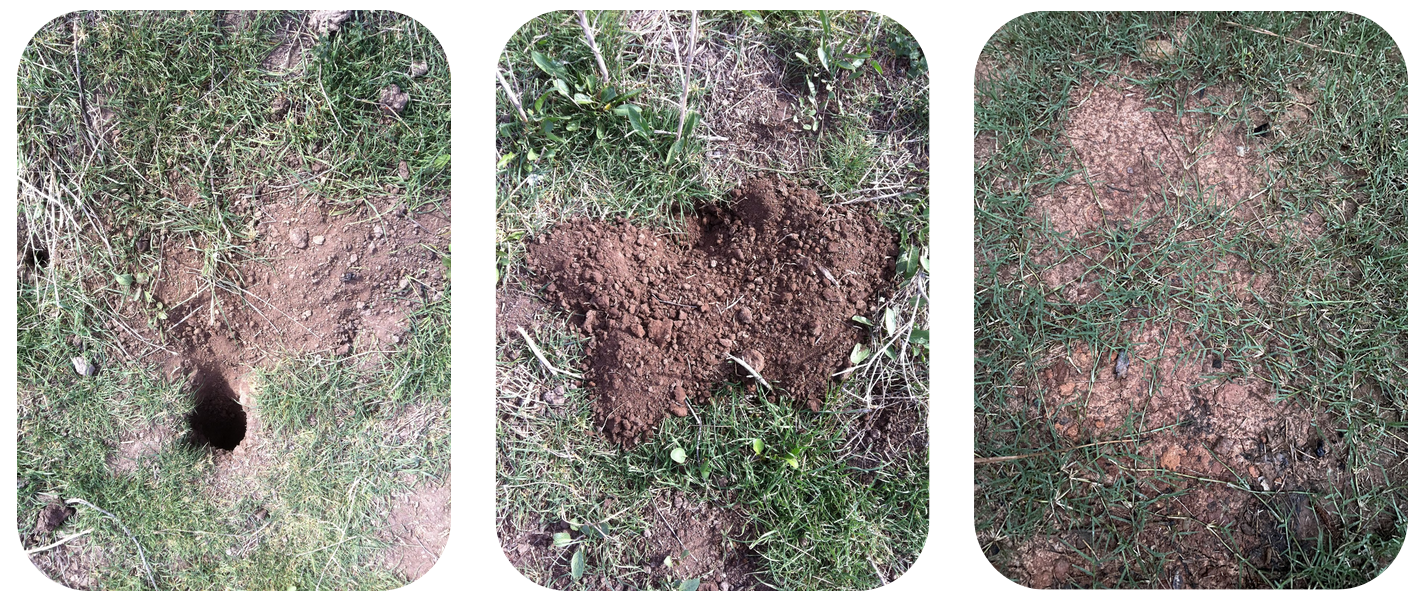}
    \caption{An illustration of the variance within the general class of ``burrow". (Left) An obvious burrow with a small flattened mound. (Center) An occluded burrow and an obvious mound; is ``mound" as important as ``hole"? (Right) A very old, flattened mound with no visible hole; how degraded a target is allowable?}
    \label{fig:hole_variety}
\end{figure}

\subsubsection{Defining Field Survey Inclusion Criteria}
\label{sec:methods_protocol_criteria}
At the outset of each field exercise in dataset construction, it is crucial to have a consistent inclusion criterion specified, so that annotators generate labels of consistent noise levels. 
See Figure~\ref{fig:hole_variety} for an example of the ambiguities of defining the training target when surveying for the class ``mammalian pest burrow".
Controlling this ambiguity through agreed-upon heuristics and ground personnel training prevents us from sending vague training signals to learning models.

\subsubsection{False Annotation Mitigation}
\label{sec:methods_protocol_mitigation}
It is critical that the drone is kept from flying through regions of the deployment environment which annotators did not cover. 
This, and also cases where annotators skipped or missed targets during their survey, generates false negatives in the resulting dataset.
In controlling these two factors from the start of the field operation, users mitigate the impact of noise in the labeling scheme and the resulting training labels.
While the flight paths of the drone are easily controlled by structured surveys, radiating from the home point of the drone, the more subtle issue of features which the annotators skipped or missed is left to Section~\ref{sec:methods_det_georef} and Appendix~\ref{app:metrics}.
In those sections, we discuss how this case leads to the failure of simple definitions of precision in geospatial feature detection.

\subsection{The BirdsEye Software Backend}
\label{sec:methods_software}

\subsubsection{Automatic Annotation Process}
\label{sec:methods_software_annotation}
We use rigid-body transforms and calibrated Brown-Conrady projective camera model for world-to-image transformations of the geotags~\cite{Hartley2004}.
The projection is dependent on high-quality pose estimates; Section~\ref{sec:bound} gives guidance on how to translate a target annotation tolerance into the pose sensor noise profile a platform must achieve to meet it.
Scalar depth \( d \) is obtained from a radar altimeter measurement.
The projective ground plane is assumed to be planar with its surface normal antiparallel to the optical axis at this distance. 
We explicitly characterize the impact of violations of this assumption on reprojection accuracy in Section~\ref{sec:results_apriltag_flatworld}.

While this geometric assumption appears drastically reductive, it is important to recognize the geometry of the application viewpoint; aerial views at altitudes defined somewhat arbitrarily by GSD requirements.
The relevant scale is therefore not the absolute relief of the scene, but the ratio of local relief to camera altitude. For the predominantly downward-looking aerial viewpoints considered here, local elevation variations constitute a relatively small perturbation to the camera-to-ground distance at the operating altitudes required by the target GSD. The resulting reprojection error is consequently governed primarily by the magnitude and spatial distribution of departures from the assumed ground plane, rather than by the existence of three-dimensional structure itself. This provides a physically meaningful basis for evaluating the approximation empirically through the coplanar-versus-noncoplanar reprojection analysis presented in Section~\ref{sec:results_apriltag_flatworld}.
Nevertheless, high-precision ray-casting method development is under way, to replace this assumption with the true world geometry and gain access to subcanopy operations or infrastructure inspection tasks.

A related question is the height of the target itself above the assumed plane.
Figure~\ref{fig:hole_variety} shows that a burrow is frequently accompanied by a soil mound, which is not a point on the ground surface. 
To first order, a target whose apex sits a height $h$ above the plane, imaged at radial distance $r$ from the optical center at altitude $d$, is displaced in projection by approximately $(h\,r/d)$. 
For the geometry used throughout this study ($d = 10$~m, $r \leq 2.5$~m at the frame corner), environmental relief of $h = 52$~cm displaces the projected point by roughly 13~cm, which is the soft-label radius of Section~\ref{sec:methods_soft_labels}. 
If this is unsustainable, the users can fly higher, balancing their GSD requirements against the impact of local relief on projection accuracy.
Target relief can therefore be absorbed by the label geometry at the operating altitudes considered here, though it becomes significant for taller targets or lower flight.

Beyond the planar scene model, the projection pipeline rests on three further assumptions, which we state explicitly because they bound where the method applies. 
First, targets are assumed static in world coordinates over the interval between the ground survey and the overflight; this holds for the burrows, structures, and persistent plant-scale features we address, and would not hold for mobile subjects. 
Second, the analysis assumes a maintained RTK fix throughout both the survey and the flight.
Across the campaigns reported here, fix was maintained under open sky; operation under canopy, near structures, or beyond the correction link, where the solution degrades to float or autonomous, falls outside the characterized envelope. 
Third, calibration parameters are treated as constant after the procedure of Section~\ref{sec:methods_calib}; we do not characterize drift over a deployment season, and a platform subject to thermal or mechanical disturbance would require periodic recalibration.

\subsubsection{Geotag Detection}
\label{sec:methods_software_geotags}

\begin{figure}
    \centering
    \includegraphics[width=\linewidth]{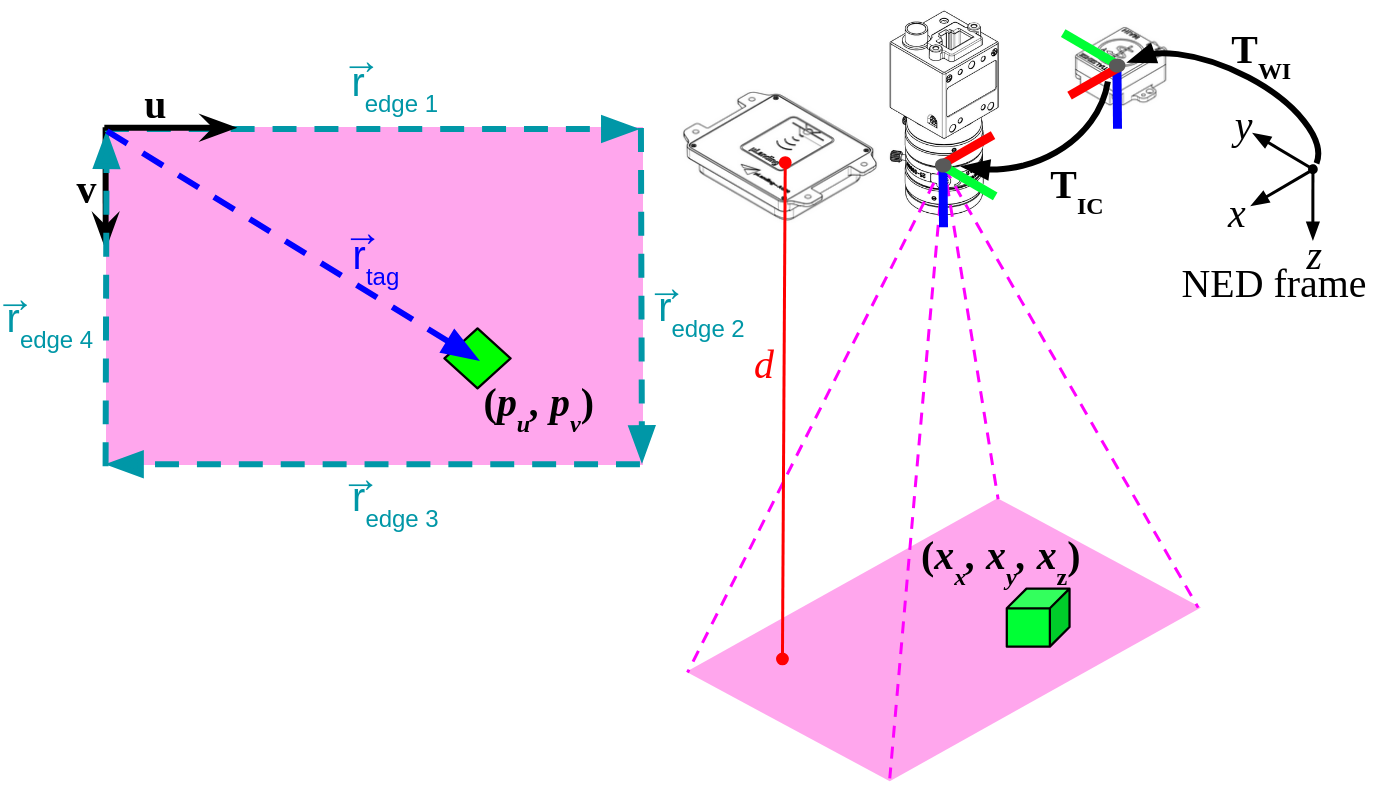}
    \caption{Illustration of the geometric relationship between a 3D geotag in the world and its 2D projection in the image. Also pictured are the position vectors relevant to the cross-products in the geotag detection method of Section~\ref{sec:methods_software_geotags}.}
    \label{fig:projection_model}
\end{figure}

To detect visible geotags, we implement the following inclusion check.
Given depth data, we project a right-angle cone to the approximated surface of the UAV's environment.
For each segment of the intersection between the cone and the approximate surface, we compute the cross product of the segment and each projected geotag's pixel coordinates. 
Let $ \overrightarrow{r}_{edge} = [x_{edge}, y_{edge}]^T $ be the displacement vector between 2 adjacent vertices of the frame boundary and $ \overrightarrow{r}_{tag} = [x_{tag}, y_{tag}]^T $ be the position vector of a projected geotag (these vectors are illustrated in Figure~\ref{fig:projection_model}).
Then:
\begin{equation}
    \label{eq:inclusion}
    \overrightarrow{r}_{edge} \times \overrightarrow{r}_{tag} = x_{edge} y_{tag} - x_{tag} y_{edge}.
\end{equation}
If this quantity is positive for a clockwise walk through all bounding rectangle segments, then the corresponding geotag is in-frame.

\subsubsection{Noise-Aware Annotations}
\label{sec:methods_soft_labels}

\begin{figure}
    \centering
    \includegraphics[width=\linewidth]{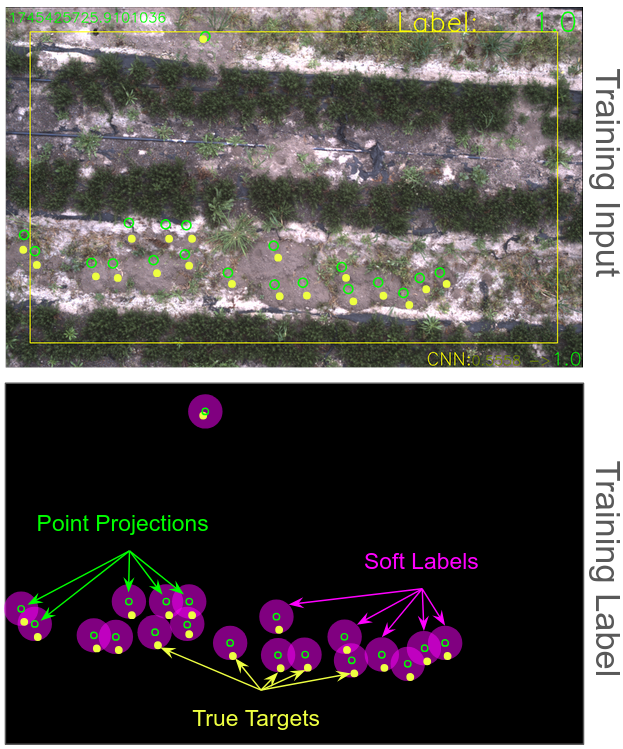}
    \caption{Our soft labeling strategy lets us use empirical or analytical projection uncertainty estimates to tune the shape of Gaussian patches, mitigating the chance that projective errors lead to truly erroneous positive annotations.}
    \label{fig:soft_labels}
\end{figure}

Labels are specified as Gaussian humps, or ``soft labels", to represent the impact of pose uncertainty on our projection (Figure~\ref{fig:soft_labels}).
These Gaussian humps are drawn out to a 3-$\sigma$ radius, with one centered on every in-frame geotag. 
The $\sigma$ parameter can be set empirically (Section~\ref{sec:results_apriltags}) or based on uncertainty propagation (Section~\ref{sec:bound}). 
In our case study, it was set to 10px, which was slightly tighter than the mean of empirical standard deviation results of Table~\ref{tab:error-norms} (13.55px).
The analytical route exists for platforms where an empirical characterization campaign is not available. 
Its predictions agree with the measured projection error of Table~\ref{tab:error-norms} at the deployed noise scale (Section~\ref{sec:results_sensitivity}, Figure~\ref{fig:scaling}), which is what licenses the substitution; the route is applied to a real system in Section~\ref{sec:results_bound}.

In practice, the radius of the soft label lets us improve certainty that true positive features produce a nonzero training signal under platform pose estimate noise.
This is in a probabilistic sense: the wider the patch is, the more sure we are that nonzero reward is placed on the target feature. This guarantee comes at the cost of diluted reward and label sparseness, where the reward placed on a target beyond 2- or 3-$\sigma$ is negligible and the number of nonzero-valued label pixels grows with patch-radius-squared, which is disadvantageous during the computation of gradients.

\subsubsection{Detection Georeferencing}
\label{sec:methods_det_georef}

Once an object has been detected in image space, its georeferencing follows naturally from the calibrated camera model. 
The detection is first back-projected through the calibrated camera to form a viewing ray or rays in the world frame. All rays are then intersected with the local planar approximation of the world surface, described in Section~\ref{sec:methods_software_annotation}. 
The resulting intersection provides the estimated world position of the detected object.
In order to quantify detection accuracy, we define recall and a precision-like quantity, which we call geotag-matched cluster fraction (GMCF), in Appendix~\ref{app:metrics}.
These metrics have pre- and post-human review definitions.

\subsection{Bounding Admissible Sensor Noise}
\label{sec:bound}

The projection error model this bound rests on is derived in Appendix~\ref{sec:methods_sensitivity}: a first-order propagation of an $\mathfrak{se}(3)$ pose perturbation through the projective function, giving $\underline{\varepsilon} \sim \mathcal{N}(0, \Sigma_u = J_\xi \Sigma_0 J_\xi^\top)$ with $J_\xi$ as in Equation~\ref{eq:jacobian_xi}. 
That linearization is checked against direct Monte Carlo perturbation of the projection pipeline in Section~\ref{sec:results_sensitivity}; the bounds below inherit its validity
regime.

\subsubsection{A Design Bound Linear in the Sensor Noise Budget}
\label{sec:methods_sensitivity_trace}

Since $\Sigma_u = J_\xi \Sigma_0 J_\xi^\top$ is $3\times3$ (rank 2, by construction) and positive semidefinite, its trace and its largest eigenvalue satisfy 
\begin{equation}
    \label{eq:trace}
    \lambda_{max}(\Sigma_u) \leq \operatorname{tr}(\Sigma_u) \leq 2\,\lambda_{max}(\Sigma_u)
\end{equation}
pointwise~\cite{HornJohnson2012}, and the trace is linear in the entries of $\Sigma_0$:
\begin{equation}
    \label{eq:trace_identity}
    \operatorname{tr}\!\left(J_\xi \Sigma_0 J_\xi^\top\right)
    = \sum_{i=1}^{6} \sigma_i^2 \, \lVert \underline{j}_i(\underline{X}_w) \rVert_2^2
    = \underline{s}^\top \underline{g}(\underline{X}_w),
\end{equation}
where $\underline{j}_i$ is the $i$-th column of $J_\xi$, $\sigma_i^2$ the corresponding diagonal entry of $\Sigma_0$, $\underline{s} = [\sigma_1^2 \; \cdots \; \sigma_6^2]^\top$ is the \emph{design vector}, and $\underline{g}(\underline{X}_w) = [\lVert \underline{j}_1 \rVert_2^2 \; \cdots \; \lVert \underline{j}_6 \rVert_2^2]^\top$ is the per-axis \emph{gain vector} at the imaged point.
The sufficient condition for the annotation tolerance to hold everywhere in frame is then:
\begin{equation}
    \label{eq:design_bound_trace}
    \boxed{\;
    \sup_{\underline{X}_w \in \mathcal{X}}\;
    \underline{s}^\top \underline{g}(\underline{X}_w)
    \;\leq\; \lambda_{u,target}\;}
\end{equation} 
The error of this estimate is bounded at a factor of two, by Equation~\ref{eq:trace}.

Equation~\ref{eq:design_bound_trace} is a half-space in $\underline{s}$: the admissible noise budgets form a convex polyhedron whose bounding hyperplane is the gain vector at the worst-imaged point.
This means that Equation~\ref{eq:design_bound_trace} is formally a semi-infinite constraint~\cite{Hettich1993}, one row per point of $\mathcal{X}$, which would ordinarily be handled by cutting-plane exchange~\cite{Kelley1960}. 
Here the supremum is attained at a single point of the frame, so the constraint reduces to one linear row per queried geometry and no exchange is required.

Another property of Equation~\ref{eq:design_bound_trace} makes it useful in practice.
The constraint functional is positively homogeneous of degree one in $\underline{s}$, so the largest uniform scaling of an existing sensor suite is available in closed form: with $\underline{s}_{base}$ the deployed budget, $\underline{s} = \kappa^\star \underline{s}_{base}$ is admissible for
\begin{equation}
    \label{eq:kappa_star}
    \kappa^\star = \frac{\lambda_{u,target}}
        {\sup_{\underline{X}_w \in \mathcal{X}}
         \underline{s}_{base}^\top \underline{g}(\underline{X}_w)}.
\end{equation}

\subsubsection{Inversion to a Per-Axis Pose Budget}
\label{sec:methods_sensitivity_budget}

Because Equation~\ref{eq:design_bound_trace} retains the individual variances, it inverts into a per-axis specification rather than a single number.
The inversion is not unique: the feasible set $\mathcal{F} = \{\underline{s} \geq 0 : \sup_{\mathcal{X}} \underline{s}^\top \underline{g} \leq \lambda_{u,target}\}$ is a six-dimensional convex body, and every point of it is a certified design, so selecting one requires a stated preference.
A linear objective such as $\max \sum_i \sigma_i^2 / \sigma_{i,base}^2$ is maximized at a vertex of $\mathcal{F}$ and places the entire budget on a single axis, which is the correct answer to that objective and not a usable specification.

We therefore report the allocation that maximizes $\sum_i \log \sigma_i^2$, the logarithm of the geometric mean of the elements of $\underline{s}$, which is strictly concave and hence maximized at a unique interior point.
On a single active row ${\underline{g}^*}^\top \underline{s} = \lambda_{u,target}$, stationarity of $\sum_i \log \sigma_i^2$ gives $1/\sigma_i^2 = \mu g^*_i$, so $g^*_i \sigma_i^2$ is the same for all $i$ and the active constraint fixes it:
\begin{equation}
    \label{eq:balanced_alloc}
    \sigma_i^2 = \frac{\lambda_{u,target}}{6\,g^*_i},
    \qquad\text{equivalently}\qquad
    g^*_i \sigma_i^2 = \frac{\lambda_{u,target}}{6}
    \quad \text{for every } i,
\end{equation}
an \emph{equal budget share}: each degree of freedom is permitted to contribute the same worst-case pixel variance, and its admissible pose uncertainty follows by dividing out the measured gain.\footnote{This is the D-optimality criterion of optimal experiment design~\cite{Pukelsheim2006} specialized to a diagonal
covariance. 
Being scale invariant, it also makes the resulting tolerances a property of the imaging geometry and $\lambda_{u,target}$ alone, independent of the deployed $\Sigma_0$; the numbers in Table~\ref{tab:design_bound_budget} are therefore a requirement on a platform rather than a description of this one.}
Both of the above forms inherit the first-order approximation of
Appendix~\ref{sec:methods_sensitivity}. Section~\ref{sec:results_sensitivity}
establishes that the linearization holds over the operational range of pose
perturbations and identifies where it begins to fail, which bounds the regime in
which these design conditions can be trusted.

\section{Results}
\label{sec:results}

\subsection{Empirical Characterization of Projection Error}
\label{sec:results_apriltags}

To quantify system projective accuracy, we designed a validation procedure using geotagged AprilTag targets in a flat, level environment. 
This setup provides both pixel-space ground truth via AprilTag detections and world-frame ground truth via geotag RTK GPS positions, enabling comprehensive evaluation of both pixel and geospatial projection fidelity.

\subsubsection{Experimental Setup and Flight Maneuvers}
\label{sec:results_apriltags_setup}

\begin{table}[ht]
    \centering
    \caption{Norms and uncertainties of error vectors from georeferencing benchmark tests.$^\dagger$}
    \label{tab:error-norms}
    \begin{tabular}{lccccc}
        \toprule
        \textbf{Altitude} & \textbf{Maneuver} & \multicolumn{2}{c}{\textbf{Error Norm}} & \multicolumn{2}{c}{\textbf{Standard Deviation}} \\
        \cmidrule(lr){3-4} \cmidrule(lr){5-6}
        & & \textbf{Meters} & \textbf{Pixels} & \textbf{Meters} & \textbf{Pixels} \\
        \midrule
        10m  & Hovering & \textbf{0.038} & 18.8 & \textbf{0.019} & 9.7 \\
                   & Yawing & 0.191 & 82.2 & 0.052 & 22.1 \\
                   & Head-on & 0.045 & 19.6 & 0.024 & 10.4 \\
                   & Strafing & 0.062 & 27.4 & 0.044 & 19.7 \\
        \midrule
        20m & Hovering & 0.043 & 9.8 & 0.037 & 8.4 \\
                   & Yawing & 0.179 & 41.1 & 0.065 & 14.8 \\
                   & Head-on & 0.081 & 18.6 & 0.033 & 7.6 \\
                   & Strafing & 0.074 & 16.9 & 0.069 & 15.7 \\
        \bottomrule
    \end{tabular}
    \begin{tablenotes}
        \item $^\dagger$\scriptsize{The data of this table gives estimates of our effective GSD at both test altitudes: 2.219$\pm$0.138\,mm @ 10\,m; 4.376$\pm$0.021\,mm @ 20\,m.}
    \end{tablenotes}
\end{table} 

We executed a series of flights at altitudes of 10\,m and 20\,m, designed to emulate common operational behaviors. 
These included static hovers and in-place yawing rotations, as well as nose-on and strafing overflights. 
Images were captured at 5\,Hz for at least 120 seconds per maneuver, yielding approximately 600 frames per maneuver iteration. 
Each maneuver was repeated five times, resulting in over 3,000 data points per maneuver category.

\subsubsection{Projection Accuracy}
\label{sec:results_apriltags_performance}

Table~\ref{tab:error-norms} summarizes the system’s projection performance using the calibration parameters deployed throughout the remainder of our experiments. 
The geospatial projection error remained within single-digit centimeters in both mean and standard deviation, excluding flights involving sustained yawing. 

\subsection{Annotation Sensitivity to Pose Uncertainty}
\label{sec:results_sensitivity}

We present a scaling study where logarithmically-spaced scale factors, between 0.1 and 100 in 25 steps, are multiplied directly by $\Sigma_0$ to observe model uncertainty evolution as the pose estimate becomes noisier.
To set up our analysis, we simulate a set of evenly-spaced test points on a flat and level ground plane (a 67 cm grid, in this study).
For the rest of this discussion, 
$$\Sigma_0 = diag(0.0001, 0.0001, 0.0036, 0.0016, 0.0016, 0.0169)$$
where $diag(.)$ creates an $n\times n$ diagonal matrix from an $n$-D vector.
Attitude entries are given in degrees² for readability and converted to radians² before propagation.

\begin{figure}[h]
    \centering
    \includegraphics[width=0.9\linewidth]{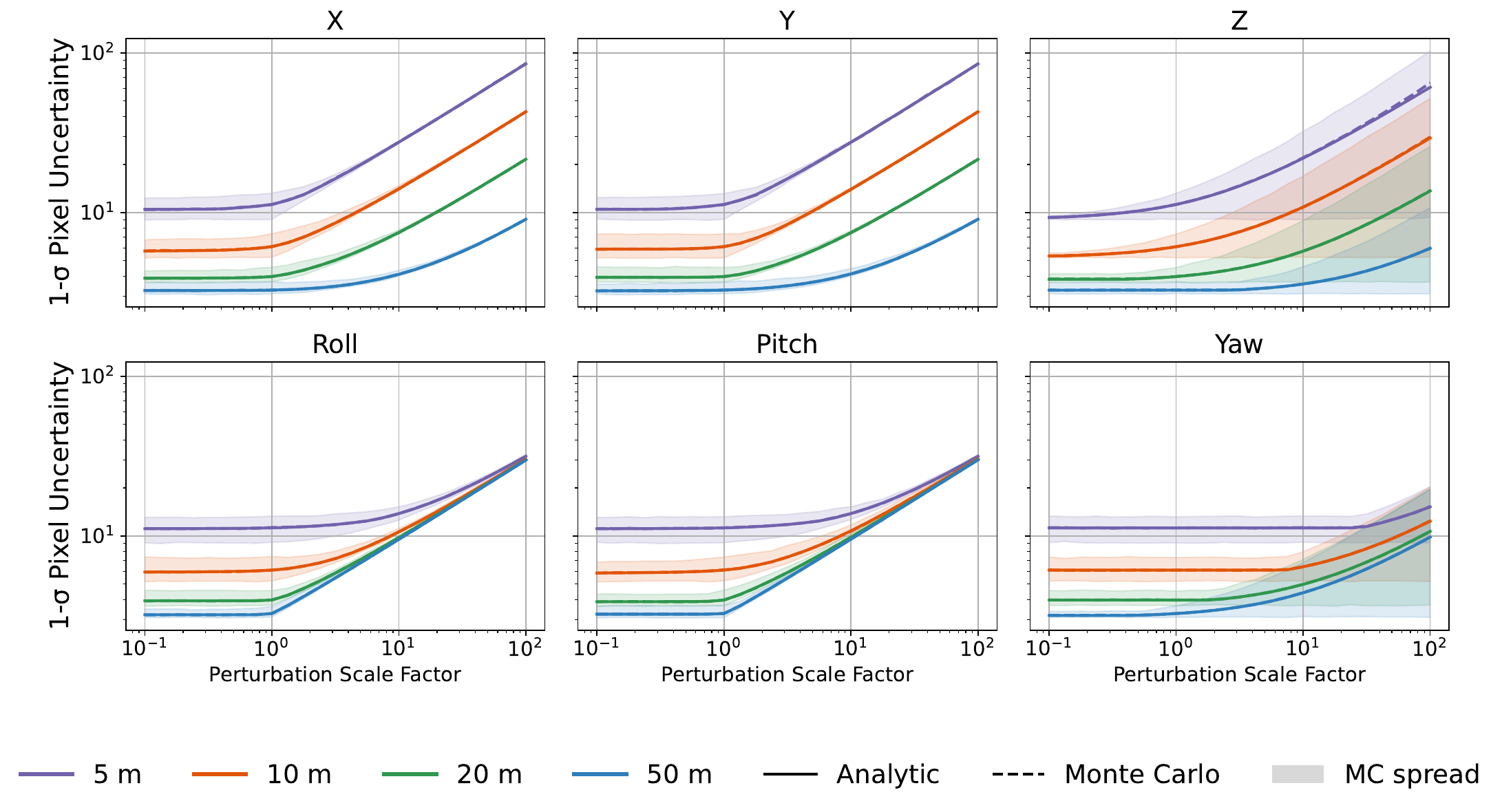}
    \caption{Perturbation study results showing pixel uncertainty sensitivity to individual pose degrees of freedom. 
    Shaded regions denote the range of Monte Carlo estimates across all visible test points at a given altitude.
    }
    \label{fig:perturbation}
\end{figure}

\begin{figure}[h]
    \centering
    \includegraphics[width=0.8\linewidth]{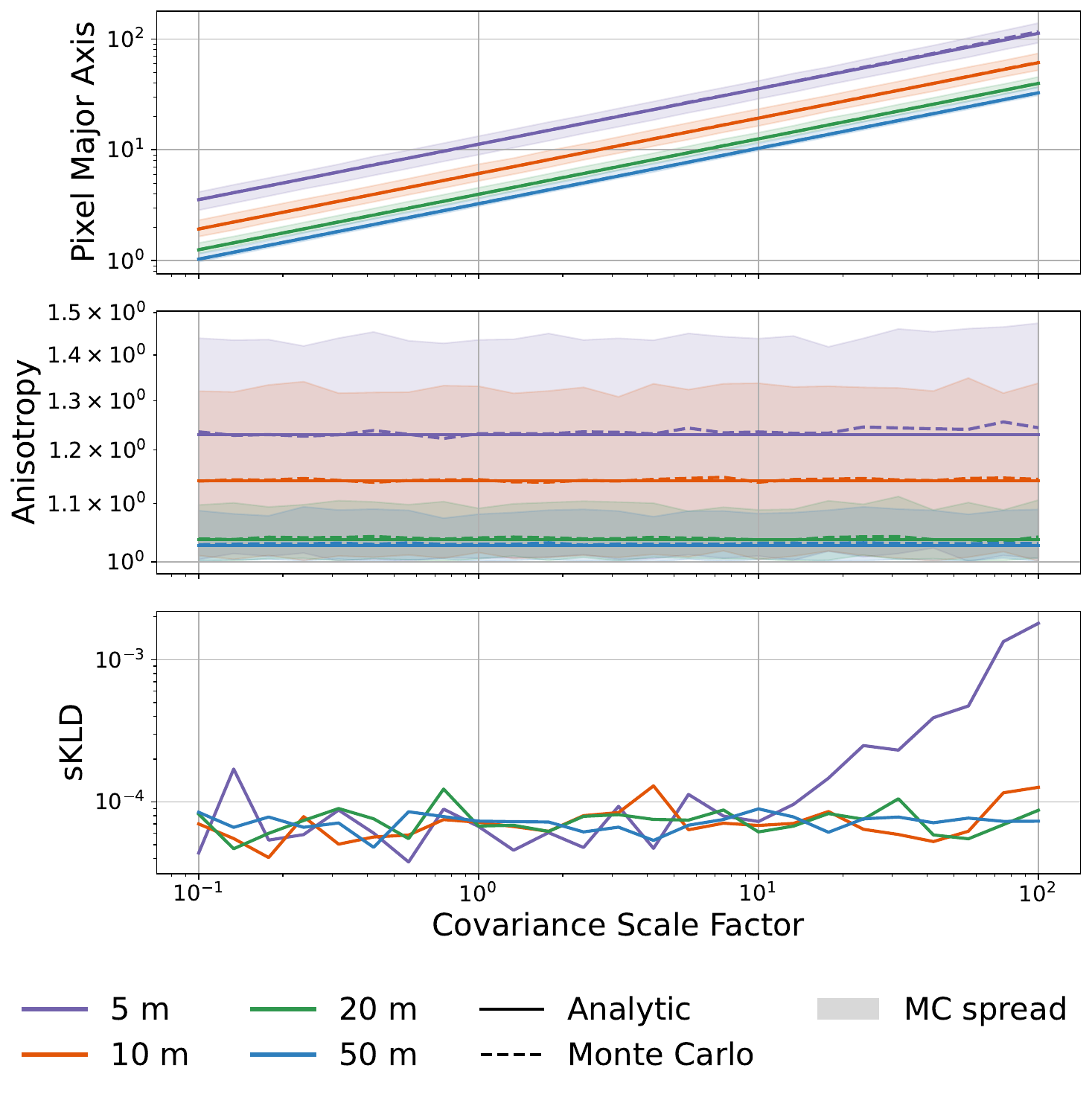}
    \caption{Scaling behavior of pixel-space uncertainty under covariance inflation. 
    \textbf{(Top)} Growth of the major axis of the uncertainty ellipse, which exhibits the expected square-root scaling law (slope $0.5$ in log--log space) predicted by first-order propagation, up to the point where linearization begins to break down. 
    \textbf{(Middle)} Evolution of ellipse anisotropy, demonstrating relative insensitivity to noise scale. 
    \textbf{(Bottom)} Similarity of the analytical model and Monte Carlo ensemble of covariance, by the symmetric Kullback-Leibler divergence (sKLD). 
    The stated heuristics for indistinguishable ($10^{-2}$) and clearly distinguishable ($10^{-1}$) distributions are well above these curves, but the 5m is beginning to diverge at scale factors above 10.
    }
    \label{fig:scaling}
\end{figure}

In Figure~\ref{fig:perturbation}, we present the scale of the major axis of the pixel uncertainty ellipse versus perturbation scale factor for the 6 degrees of freedom in a pose estimate.
Here, we see that the system is insensitive to perturbations in low-noise pose parameters until the scaling inflates their magnitude above the preceding largest diagonal element.
Altitude uncertainty is our dominant contributor to error, especially at low altitudes.
At higher scale factors, we see that the higher altitude flights grow more sensitive to angular pose parameter noise, though this inflation is small relative to the maximum major axis scale among all test points.

In Figure~\ref{fig:scaling}, we see that our analytical model agrees with the empirical results of Section~\ref{sec:results_apriltags_performance} (Table~\ref{tab:error-norms}) for scale factors around unity, where the computed major axis of the uncertainty ellipse is roughly 15px or less for all altitudes simulated (with larger magnitude on lower flights).
This empirically supports the theoretical analysis applied.
The bottom panel quantifies the divergence between analytical and Monte Carlo predictions using the symmetric Kullback–Leibler divergence (sKLD)\footnote{Jeffreys defined this divergence several years prior to its appearance in Kullback and Leibler's work, so it is also known as the Jeffreys divergence.}.
We see strong agreement between the Monte Carlo validation results and the analytical results, where the only divergence we see is at scale factors above 10 at 5m AGL flight.
Heuristically, sKLD values less than $10^{-2}$ are practically indistinguishable, while values in $(10^{-2},\;10^{-1}]$ entail that the models are diverging and values above $10^{-1}$ mean that the models are distinctly different.
These thresholds are at least one order of magnitude above the curves of Figure~\ref{fig:scaling}. 

\subsection{Applying the Sensor Noise Analysis}
\label{sec:results_bound}

Table~\ref{tab:design_bound_variants} reports $\kappa^\star$ for Equation~\ref{eq:design_bound_trace} and for the exact worst-case reference $\sup_{\mathcal{X}} \lambda_{max}(\Sigma_u)$, which is also linear in $\underline{s}$ for a fixed maximizing direction and therefore admits the same treatment at the cost of an eigendecomposition per query.

\begin{table}[ht]
    \caption{Admissible uniform scaling $\kappa^\star$ of the deployed pose-noise budget (Equation~\ref{eq:kappa_star}) under two criteria, at
    $\lambda_{u,target} = \bsLamTarget$\,px$^2$ ($r = 30$\,px at $3\sigma$).
    $f(\underline{s}_{base})$ is the bound's estimate of the worst-case pixel
    variance at the deployed $\Sigma_0$; the true value is
    $\sup_{\mathcal{X}}\lambda_{max}(\Sigma_u) = \bsSupLmaxTen$\,px$^2$ at $h = 10$\,m and $\bsSupLmaxTwenty$\,px$^2$ at $h = 20$\,m.
    The trace condition is an upper bound on $\lambda_{u,max}$ and the eigenvalue condition is exact. The ``penalty'' column shows $\kappa^\star$(exact)$/\kappa^\star$(row).
    Suprema are evaluated over 2,307,121 sampled points per altitude at 1 px stride.
    }
    \label{tab:design_bound_variants}
    \centering
    {\small\setlength{\tabcolsep}{4pt}
\begin{tabular}{llrrrl}
\toprule
\textbf{Sufficient condition} & \textbf{AGL} & \textbf{$f(\underline{s}_{base})$ [px$^2$]} & \textbf{$\kappa^\star$} & \textbf{Penalty} & \textbf{Verdict} \\
\midrule
Eq.~\ref{eq:design_bound_trace} (trace)   & 10\,m & \bsSupTraceTen    & \bsKappaTraceTen       & \bsGapTen$\times$       & reject ($1.09\times$) \\
                                          & 20\,m & 46.42             & \bsKappaTraceTwenty    & \bsGapTwenty$\times$    & accept \\
\midrule
$\sup_{\mathcal{X}}\lambda_{max}(\Sigma_u)$ (exact) & 10\,m & \bsSupLmaxTen & \bsKappaExactTen & 1.000$\times$      & accept \\
                                          & 20\,m & \bsSupLmaxTwenty  & 3.794                  & 1.000$\times$           & accept \\
\bottomrule
\end{tabular}
}
\end{table}

At this operating point the residual factor-of-two slack is decision-relevant.
Equation~\ref{eq:design_bound_trace} gives $\kappa^\star = \bsKappaTraceTen$ at
$h = 10$\,m---a rejection by $8\%$---whereas the exact constraint gives
$\kappa^\star = \bsKappaExactTen$, an acceptance with
$\bsKappaInvTen\times$ headroom.
Both are valid upper bounds; the trace form is simply not tight enough to
certify hardware that sits this close to the tolerance boundary, but the
eigenvalue form is.
At $h = 20$\,m, where $\kappa^\star = \bsKappaTraceTwenty$ under the trace
bound, either form answers the question.
The practical recommendation is therefore to use
Equation~\ref{eq:design_bound_trace} for allocation and screening, where its
linearity in $\underline{s}$ is the point, and to confirm a near-boundary
platform against $\sup_{\mathcal{X}} \lambda_{max}$.

Table~\ref{tab:design_bound_budget} is a design target, not a per-axis
pass/fail test: conformance is decided by Equation~\ref{eq:design_bound_trace}
evaluated at the deployed budget, which the platform satisfies, and an axis may
exceed its balanced allocation while the platform as a whole conforms.

\begin{table}[ht]
    \caption{Admissible per-axis pose uncertainty ($1\sigma$) required to hold the
    soft-label radius used in training ($r = 30$\,px at $3\sigma$,
    $\lambda_{u,target} = \bsLamTarget$\,px$^2$) at every point in frame, from
    Equation~\ref{eq:design_bound_trace} with the balanced allocation of
    Section~\ref{sec:methods_sensitivity_budget}.
    Deployed values are the datasheet $1\sigma$ figures of
    Section~\ref{sec:results_sensitivity}.
    Bold entries are exceeded by the deployed platform.
    Admissible translation tolerances scale linearly with altitude; rotation
    tolerances are altitude-invariant at nadir. The final column is each axis's
    share of the deployed platform's total worst-case pixel variance,
    $g^\star_i\sigma_{i,\mathrm{dep}}^2 / \sum_j g^\star_j\sigma_{j,\mathrm{dep}}^2$;
    it describes the deployed $\Sigma_0$ and is not the equal $1/6$ share that the
    balanced allocation assigns by construction. Altitude accounts for
    44.7\% of that total, which is why $z$ is the axis over its allocation.}
    \label{tab:design_bound_budget}
    \centering
    {\small\setlength{\tabcolsep}{4pt}
\begin{tabular}{lrrrc}
\toprule
\textbf{Axis} & \textbf{Deployed $1\sigma$} & \textbf{Balanced, 10\,m} & \textbf{Balanced, 20\,m} & \textbf{Budget Share at $s_{base}$, 10\,m} \\
\midrule
$x$      & 10\,mm            & 13.64\,mm          & 27.28\,mm           & 16.7\% \\
$y$      & 10\,mm            & 13.45\,mm          & 26.89\,mm           & 16.7\% \\
$z$      & 60\,mm            & \bsSigmaZTen\,mm   & \bsSigmaZTwenty\,mm & \textbf{42.9\%} \\
roll     & 0.04$^{\circ}$   & 0.0753$^{\circ}$   & 0.0753$^{\circ}$    & 8.5\% \\
pitch    & 0.04$^{\circ}$   & 0.0741$^{\circ}$   & 0.0741$^{\circ}$    & 9.0\% \\
yaw      & 0.13$^{\circ}$   & \bsSigmaYawTen$^{\circ}$ & \bsSigmaYawTen$^{\circ}$ & 6.1\% \\
\bottomrule
\end{tabular}
}
\end{table}

\subsection{Annotation Sensitivity to Terrain Slope}
\label{sec:results_apriltag_flatworld}

The BirdsEye annotation pipeline currently assumes visible terrain is coplanar with the floor of the projective cone, using a single depth estimate per frame derived from radar altimetry; we do not assume that the world is globally planar, but rather that the ground within view in a given frame can be reasonably approximated by a plane.
To quantify the robustness of this assumption, we conducted a controlled simulation study to analyze error propagation under systematic violations of coplanarity.
Assuming that the ground beneath the drone has low curvature and undulation within the field of view, then two factors dominate the contribution of terrain slope to projective error: (i.) the angle between the normal vectors of the ground plane and the normal vector of the projective cone's floor and (ii.) the distance along the slope a point is from the optical center.

\begin{figure}[h]
    \centering
    \includegraphics[width=\linewidth]{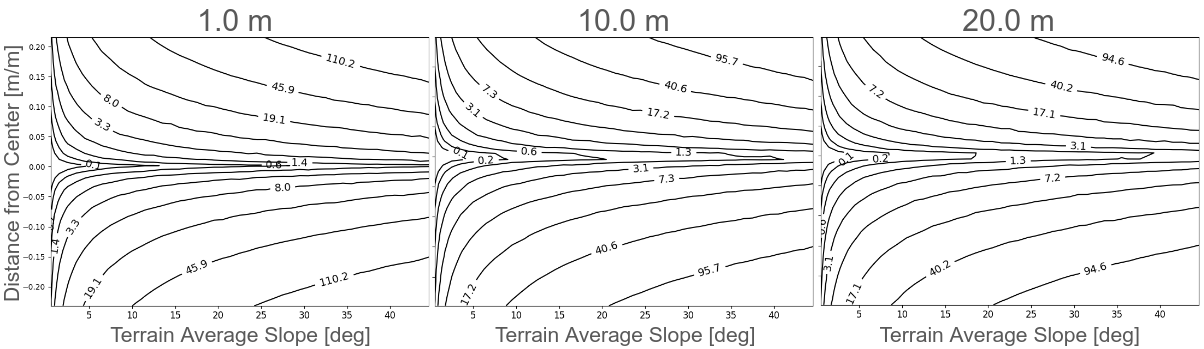}
    \caption{Error propagation under violations of the annotation projection pipeline flat-world assumption.  
    The horizontal axis represents the terrain slope angle and the vertical axis represents the distance, in meters, of a test point from the optical center, normalized by the altitude, in meters, of the camera and contour labels are in units of pixels.
    We note that terrain-induced error exceeds the magnitude of pose-induced error at slopes of approximately 10 degrees.}
    \label{fig:error_prop}
\end{figure}

We evaluated reprojection error as a function of these two variables by densely sampling the camera field of view and constructing corresponding test points on both the true sloped surface and the approximated planar surface used by the annotation pipeline.
Terrain slopes from 0$^{\circ}$ to 45$^{\circ}$ were evaluated in 0.5$^{\circ}$ increments.
The results are displayed in Figure~\ref{fig:error_prop}. 
Comparison with the reprojection error induced by flight maneuvers (Table~\ref{tab:error-norms}) shows that terrain-induced error exceeds pose-induced error at average slopes of approximately 10$^{\circ}$.

In flat-and-level flight, reprojection error therefore remains within the same order of magnitude as motion-induced error for terrain slopes up to roughly 10$^{\circ}$, which we take as the nominal operating envelope of the flat-world approximation.
The system’s effective ground sample distance (2.2 mm at 10 m AGL and 4.4 mm at 20 m AGL) yields georeferencing errors of approximately 8 cm at 10 m AGL and 18 cm at 20 m AGL. 
Beyond 10$^{\circ}$ slopes, error increases rapidly (Figure~\ref{fig:error_prop}), motivating future integration of terrain reconstruction methods for steeper and nonplanar environments.
Since our present focus is predominantly agricultural, we note that this slope envelope is broader than the operational capabilities of typical mechanized agricultural equipment, where slopes with grades beyond 13\% (7.41$^\circ$) are considered to not allow their use~\cite{Gasparini2026Apr}.

\begin{table}
    \centering
    \caption{(Upper) Summary of principal flight campaign. (Middle) Summary of subsequent field timing experiments performed at the Campus Farm. (Lower) Comparison of annotation strategy productivity. All flights performed at nominally 10m AGL. The ``Dataset Role" columns show which flights were partitioned into the training, validation, and test sets of the standard ML training pipeline.}
    \label{tab:valflight-summary}
    \begin{tabular}{lcccc}
        \toprule
         \textbf{Location} & \multicolumn{3}{c}{\textbf{Dataset Composition}} & \textbf{Dataset Role} \\
        \cmidrule(lr){2-4}
        & \textbf{\# Geotags} & \textbf{\# Frames} & \textbf{Annotations} & \\
        \midrule
        Campus Farm & 1,181 & 10,034 & 15,000 & \textit{Train}\\
        Campus Farm & 308 & 6,514 & 10,578 & \textit{Train} \\
        Partner Site 1 & 161 & 2,120 & 6,354 & \textit{Test} \\
        Campus Farm & 256 & 3,371 & 8,117 & \textit{Validation} \\
        Partner Site 2 & 181 & 2,555 & 6,613 & \textit{Train} \\
        Campus Farm & 151 & 3,647 & 11,715 & \textit{Validation} \\
        Partner Site 2 & 390 & 5,440 & 8,490 & \textit{Train} \\
        \midrule
        \textit{\textbf{Totals}} & \textit{2,628} & \textit{33,681} & \textit{66,867} & \\
        \bottomrule
    \end{tabular}
    \\[0.5em]
    \begin{tabular}{lcccccc}
        \toprule
        \multicolumn{3}{c}{\textbf{Dataset Composition}} & \multicolumn{3}{c}{\textbf{Dataset Build Time}} & \textbf{Dataset Role} \\
        \cmidrule(lr){1-6}
        \textbf{Geotags} & \textbf{Frames} & \textbf{Labels} & \textbf{Field} & \textbf{Post} & \textbf{Total} & \\
        \midrule
        999 & 3,976 & 25,315 & 4:14:34 & 0:14:27 & 4:29:01 & \textit{Train}\\
        1,015 & 4,910 & 10,573 & 3:29:06 & 0:19:29 & 3:48:35 & \textit{Validation} \\
        1,303 & 3,638 & 19,712 & 3:38:54 & 0:12:06 & 3:51:00 & \textit{Train} \\
        \midrule
        \textit{3,317} & \textit{12,524} & \textit{55,600} & 11:22:34 &  & \textit{12:08:36} & \\
        \bottomrule
    \end{tabular}
    \\[0.5em]
    \begin{tabular}{lcccc}
        \toprule
         & \textbf{Workers} (W) & \textbf{Frames} (F) & \textbf{Time} (HH:MM:SS) & \textbf{Rate} (F/s/W) \\
        \midrule
        Manual  & 15 & 1,500 & 04:48:52 & 0.006 \\
        \textit{BirdsEye} & 2 & 12,524 & 11:22:34 & 0.153 \\
        \bottomrule
    \end{tabular}
\end{table}
 
\subsection{Deployment Case Study: Burrowing Mammals}
\label{sec:results_case}

To evaluate the efficiency of the annotation pipeline, 2 field workers conducted a field campaign at three sites within a 25 mile radius (our on-campus farm and 2 local farms). 
Table~\ref{tab:valflight-summary} summarizes the primary dataset campaign, the secondary timing experiment campaign, and resulting dataset compositions.
One of the partnering farms (Partner Site 1) was held out as an unseen test dataset; Tables~\ref{tab:holdout} and ~\ref{tab:adjudication} present detection results on this geographically distinct and unseen environment.

Annotation rate comparisons against 15 non-specialist annotators, including the original 2 field workers, each tasked with annotating 100 random frames from our dataset using GUI-based methods, are shown in the bottom panel of Table~\ref{tab:valflight-summary}. 
The original field workers instructed the other participants in this study on what they considered a valid target during the field work before the experiment commenced.
This comparison shows that \textit{BirdsEye} achieves an averaged 25.5$\times$ increase in per-worker annotation rate.
Note that these productivity estimates are functions of camera framerate; they can be changed arbitrarily by adjusting capture framerates; in this discussion, the camera was capturing at 5Hz.
A framerate-invariant statistic for worker productivity can also be drawn from the bottom block of Table~\ref{tab:valflight-summary}.
The average number of unique targets identified (geotags) per worker-hour is roughly 146.
A corresponding figure in the pixel annotation domain is diluted by the repetition of a given target over the frames which see it.

\subsection{Held-Out Evaluation: Does the Dataset Transfer?}
\label{sec:heldout}

The detector in this section is an instrument, not a contribution. Its only purpose is to
test whether a dataset built by the \textit{BirdsEye} workflow is good enough to train a
model that works at a site never used to build or tune anything. No architecture comparison
is claimed, no checkpoint is recommended, and the numbers reported below are the ones a rule
selected before the test data were scored, including where they are unflattering.

\subsubsection{The Detector and Training Protocols}
\label{sec:detector}
We parse our annotated frames to 224x224 subtiles, which places greater emphasis on small-scale, local features. 
Labels are specified as isotropic Gaussian humps, per Section~\ref{sec:methods_soft_labels}. 
In practice, our soft label patches measure 60 pixels (13cm at 10m AGL) in diameter.

The backbone of our detector is an EfficientNetV2-B3. 
A stack of $3\times3$ transposed convolutions decodes the feature map back to input resolution and a final $3\times3$ convolution emits a single-channel detection heatmap. 
Training proceeds in two stages: the decoder is trained to an early stopping criterion with the backbone frozen, then the top 30\% of backbone layers are unfrozen and training restarts. 
We use the AdamW optimizer at learning rate $1\times10^{-5}$ with $\beta_1 = 0.9$ and $\beta_2 = 0.999$, standard binary cross-entropy loss, and stop after 15 epochs without improvement in validation loss. 
The negative class dominates the case study dataset, so frames containing no positive example are randomly subsampled. 
We trained 30 models of this architecture, varying preprocessing jitter settings, filter counts in the upsampling layers, and output activation.
Then we performed a hyperparameter sweep across: (i.) the raw heatmap confidence threshold, used to define detection events; (ii.) $\varepsilon$, the DBSCAN radius used to cluster detection events spatially; (iii.) $m$, the DBSCAN minimum density used to spatially cluster detection events; and (iv.) $\varepsilon_{pred}$, the prediction-mode spatial clustering radius.
All model checkpoints at all operating points in the swept hyperparameter field were screened on three validation flights, leading to the selection of the 12 presented (checkpoint, operating point) pairs at the PR-frontier.

The screening procedure, the two degeneracy guards it applies, and the per-checkpoint results are given in Appendix~\ref{app:screen}.
Additionally, the appendix describes our pre-registration procedure which pinned \bsPreregArms{} confirmatory arms (Table~\ref{tab:arms}), each fixing both a checkpoint and its complete operating point based upon validation performance, prior to revealing the held-out test set to any model.

\subsubsection{Held-out Results}
\label{sec:holdout}

\begin{figure}[ht]
    \centering
    \includegraphics[width=\linewidth]{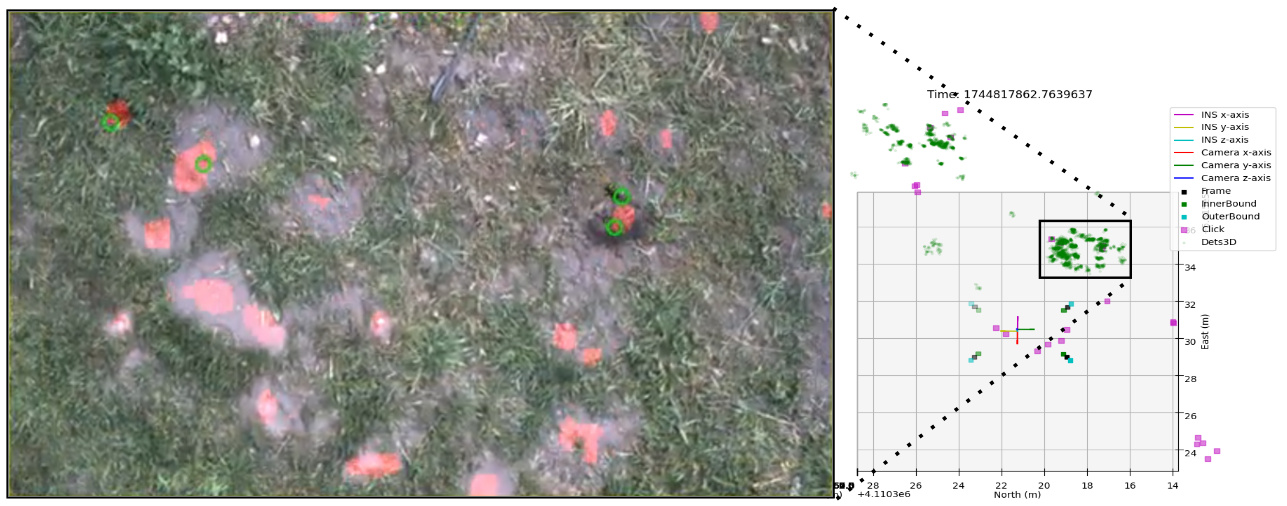}
    \caption{A georeferenced image frame from the test flight, showing both human-made field annotations (Green Rings) and CNN detections (Red Patches). Observe that the model detects both the concrete, positive training examples established by human experts and also borderline cases. This difference in behavior, where humans isolate good training examples and CNNs indicate every detection, leads to the failure of the simple precision statistic as a performance assessment in our case study. To the right of the image frame is a georeferenced detection map under construction, where magenta squares represent human field annotations and green dots represent the centroids of model detection patches.}
    \label{fig:detections}
\end{figure}

On the unseen site recall \emph{rose} and \bsGmcfName{} \emph{fell}. 
Across all \bsPreregConfigs{} scored configurations the mean change from validation to held-out is $\bsMeanDRecall$ in recall and $\bsMeanDPrec$ in \bsGmcfName, and the rank correlation between validation F1 and held-out F1 is $\bsSpearman$ -- that is, the selection-only F1 carries essentially no information about held-out F1 ordering. 

The mechanism is measurement, not model. The validation flights are surveyed \bsSurveyDensityRatio$\times$ more densely than the test flight (\bsValDensity{} against \bsTestDensity{} in-view geotags per frame), and \bsGmcfName{} is positive-unlabeled with a density-dependent bias (Appendix~\ref{app:metrics}). 
Ultimately, this incompatibility means that models are penalized more strongly for sensitivity when the target survey is sparse in a flight environment.
Therefore \bsGmcfName{}, and any F1 built on it, is not comparable across sites of different survey density. 
Recall is transferable, because its denominator is the survey unit itself (geotags). 


\begin{table}[h]
  \caption{The \bsPreregArms{} pre-registered confirmatory arms. Each arm fixes
  a checkpoint and a complete operating point
  ($t$~/~$\varepsilon$~/~$m$~/~$\varepsilon_{\mathrm{pred}}$) from the
  exact-inference pooled sweep. Validation figures are the micro-averaged
  screening values that the rule selected on; F1 is the selection-only
  quantity.}
  \label{tab:arms}
\begin{tabularx}{\textwidth}{lccCCC}
\toprule
Arm & Ckpt & $t$ / $\varepsilon$ / $m$ / $\varepsilon_{\mathrm{pred}}$ & Recall &  \bsGmcfName{} & F1 \\
\midrule
Max pooled F1 & 014 & 0.08 / 0.2 / 2 / 0.2 & 0.466 & 0.480 & 0.472 \\
Max pooled recall & 030 & 0.04 / 0.1 / 2 / 0.2 & 0.805 & 0.148 & 0.250 \\
Lowest review burden & 013 & 0.035 / 0.2 / 2 / 0.2 & 0.434 & 0.417 & 0.425 \\
\bottomrule
\end{tabularx}

\end{table}

\begin{table}[h]
  \centering
  \caption{Held-out PRE statistics on the unseen site, whole flight, \bsTestFrames{} frames processed (3 of the 2120 had duplicated timestamps), \bsTestInView{} of \bsTestGeotags{} geotags in view under Equation~\ref{eq:inclusion}. 
  Recall is over in-view geotags. The top 3 rows are the pre-registered checkpoint configurations, while the bottom 9 are the remainder of the field.}
  \label{tab:holdout}
\begin{tabularx}{0.6\textwidth}{lccc}
\toprule
Ckpt & $t$ / $\varepsilon$ / $m$ / $\varepsilon_{\mathrm{pred}}$ & $R_{\mathrm{PRE}}$ & \bsGmcfName{}$_{PRE}$ \\
\midrule
013 & 0.035 / 0.2 / 2 / 0.2 & 0.583 & 0.189 \\
014 & 0.08 / 0.2 / 2 / 0.2 & 0.563 & 0.223 \\
030 & 0.04 / 0.1 / 2 / 0.2 & 0.887 & 0.058 \\
\midrule
015 & 0.08 / 0.1 / 2 / 0.35 & 0.815 & 0.173 \\
017 & 0.08 / 0.15 / 5 / 0.2 & 0.695 & 0.235 \\
021 & 0.12 / 0.15 / 5 / 0.2 & 0.702 & 0.164 \\
028 & 0.12 / 0.1 / 20 / 0.35 & 0.669 & 0.315 \\
030 & 0.06 / 0.15 / 5 / 0.2 & 0.762 & 0.198 \\
031 & 0.08 / 0.15 / 5 / 0.2 & 0.682 & 0.235 \\
042 & 0.06 / 0.15 / 2 / 0.2 & 0.649 & 0.185 \\
044 & 0.03 / 0.15 / 10 / 0.2 & 0.689 & 0.217 \\
045 & 0.05 / 0.2 / 2 / 0.2 & 0.689 & 0.178 \\
\bottomrule
\end{tabularx}

\end{table}

\begin{table}[h]
  \centering
  \caption{Adjudicated GMCF and recall per pre-registered arm, at the
  parameters the review was conducted at ($\varepsilon = \bsReviewEps$~m,
  $m = \bsReviewMinSamples$, $\varepsilon_{\mathrm{pred}} =
  \bsReviewMatchRadius$~m, \bsReviewDedupCm~cm grid dedup). 
  The PRE column here is therefore \emph{not} the pre-registered geometry of Table~\ref{tab:holdout}. 
  POST columns are given under all three tie-break conventions in the order positive-wins / majority-wins / hard-negative-wins.
  Every POST recall is an upper bound and is printed with $\geq$.
  The top 3 rows are the pre-registered checkpoint configurations, while the bottom 9 are the remainder of the field.}
  \label{tab:adjudication}
{\small\setlength{\tabcolsep}{4pt}
\begin{tabular}{@{}l r r r r r r r r r @{}}
\toprule
& & & & \multicolumn{3}{c}{GMCF$_{POST}$} & \multicolumn{3}{c}{R$_{POST}$ ($\geq$)} \\
\cmidrule(lr){5-7}\cmidrule(lr){8-10}
Ckpt & Clusters & GMCF$_{PRE}$ & R$_{PRE}$ & pos & maj & neg & pos & maj & neg \\
\midrule
013 & 132 & 0.432 & 0.490 & 0.780 & 0.742 & 0.705 & 0.615 & 0.609 & 0.603 \\
\textbf{014} & \textbf{113} & \textbf{0.522} & \textbf{0.536} & \textbf{0.867} & \textbf{0.832} & \textbf{0.832} & \textbf{0.634} & \textbf{0.630} & \textbf{0.630} \\
030 (t=0.04) & 445 & 0.189 & 0.868 & 0.404 & 0.339 & 0.279 & 0.922 & 0.914 & 0.909 \\
\midrule
015 & 201 & 0.373 & 0.682 & 0.726 & 0.682 & 0.632 & 0.787 & 0.780 & 0.773 \\
017 & 225 & 0.324 & 0.715 & 0.684 & 0.627 & 0.578 & 0.819 & 0.809 & 0.803 \\
021 & 295 & 0.247 & 0.742 & 0.539 & 0.478 & 0.417 & 0.839 & 0.830 & 0.822 \\
028 & 382 & 0.202 & 0.801 & 0.455 & 0.393 & 0.332 & 0.882 & 0.872 & 0.865 \\
030 (t=0.06) & 269 & 0.279 & 0.735 & 0.617 & 0.558 & 0.509 & 0.838 & 0.830 & 0.824 \\
031 & 215 & 0.349 & 0.728 & 0.698 & 0.647 & 0.595 & 0.823 & 0.815 & 0.809 \\
042 & 127 & 0.409 & 0.497 & 0.819 & 0.780 & 0.748 & 0.627 & 0.620 & 0.616 \\
044 & 294 & 0.231 & 0.742 & 0.466 & 0.384 & 0.323 & 0.828 & 0.815 & 0.805 \\
045 & 192 & 0.396 & 0.656 & 0.750 & 0.693 & 0.635 & 0.767 & 0.758 & 0.749 \\
\bottomrule
\end{tabular}
}

\end{table}

\begin{figure}[h]
  \centering
  \includegraphics[width=\linewidth]{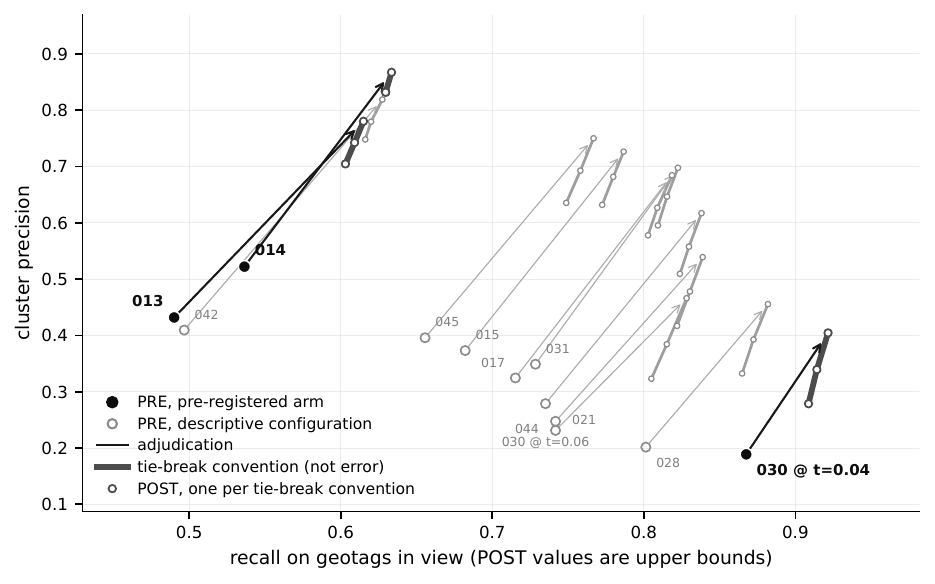}
  \caption{Effect of adjudication on every reviewed checkpoint layer, each at
  the review's parameters ($\varepsilon = \bsReviewEps$~m,
  $m = \bsReviewMinSamples$, $\varepsilon_{\mathrm{pred}} =
  \bsReviewMatchRadius$~m). Filled and open circles are the PRE
  (geometry-only) points; arrows run to the adjudicated result; the thick
  segment spans the three tie-break conventions and carries one white-filled
  marker per convention, because there are exactly three discrete outcomes
  there and not a continuum. The segment is a convention, not an error bar. The
  three layers carrying pre-registered arms are drawn dark and labeled in
  bold: \bsArmMaxFOneCkpt{} (max F1),
  \bsArmLowestBurdenCkpt{} (lowest review burden) and \bsArmMaxRecallCkpt{} at
  $t = \bsArmMaxRecallThresh$ (max recall); the remaining layers are drawn
  light. The whole reviewed field is shown because it is the evidence that the
  arms are ordinary members of it rather than the points where adjudication
  happened to help: every layer moves up and to the right, and the largest
  precision gain belongs to a layer no arm selected. POST recall values are
  upper bounds ($a = \bsAdjAddedUnits$).}
  \label{fig:prepost}
\end{figure}

\subsubsection{Adjudication}
\label{sec:adjudication}
Detections that no geotag matched were reviewed in a purpose-built interactive tool. 
For review, matching was per-point and many-to-one: a detection counted as matched if any geotag fell within the spatial threshold, and a geotag counted as found if any detection did likewise. 
Duplicate detections of a single real target therefore did not enter the review pool for reasons unrelated to model error. 
Remaining unmatched detections were grouped per frame by complete-linkage clustering at a 30~px radius, bounding every within-cluster pairwise distance.

The review unit was the consensus cluster across all checkpoint layers rather than one cluster per model. 
Every detection of every screened checkpoint on the test flight was therefore reviewed in a single pass over their union: the reviewer judged each physical spot once without reference to which model produced it, the record stored which checkpoints fired there, and every checkpoint's scorecard was derived from that one pass. 
No checkpoint was reviewed under a different standard than another, and degenerate checkpoints contributed no review targets while retaining their own full scorecards. 
Each cluster received one of four verdicts (\texttt{confirmed\_real}, \texttt{hard\_negative}, \texttt{uncertain}, or \texttt{unreviewed})) with the latter two treated as abstentions. 
Reviewers could also mark targets the detector never fired on; these enter the accounting solely as recall-denominator terms, but none were placed in this adjudication.

A verdict could optionally propagate to the same physical spot in other frames through UTM DBSCAN clusters ($\varepsilon = 0.20$~m), which span frames by construction. 
Because chaining at that radius can merge physically distinct spots, clusters holding both a positive and a negative verdict are resolved under an explicitly named convention. 
Propagation defaults to off and no convention is privileged: the choice is worth roughly 15 precision points on this dataset, so all three are computed and reported, and the spread between them is treated as the uncertainty rather than resolved away. 
Every judgment was logged with its provenance (judged directly, or propagated under a named rule) together with the matching parameters, per-checkpoint post-processing configuration, and input digests in effect when it was made.

In total, \bsAdjTotal{} adjudications, of which \bsAdjDirect{} (\bsAdjDirectPct\%) were made directly on the cluster union (\bsAdjDirectConfirmed{} confirmed, \bsAdjDirectHardNegative{} rejected) and \bsAdjPropagated{} were propagated to another frame's view of the same physical spot under the \texttt{\bsAdjPropRule{}} convention. 
These numbers are reported \emph{at the parameters the review was conducted at} ($\varepsilon = \bsReviewEps$~m, $m = \bsReviewMinSamples$, $\varepsilon_{\mathrm{pred}} = \bsReviewMatchRadius$~m) and not at the pre-registered operating points of Appendix~\ref{app:prereg}. 
Table~\ref{tab:adjudication} consequently reports PRE and POST at the review's parameters, the pre-registered geometry for the same checkpoints is given in Table~\ref{tab:holdout}; their PRE columns are not the same quantity.

Adjudication raises both scores on every detector configuration, and the tie-break convention sets how far. 
On the max-F1 arm GMCF$_{PRE}$ \bsArmMaxFOneGeoPre{} becomes \bsArmMaxFOnePostNeg{} to \bsArmMaxFOnePostPos{} depending on the convention (Table~\ref{tab:adjudication}, Figure~\ref{fig:prepost}). 
No single POST value is reported anywhere in this paper, because the choice among the three is made by convention and not by data, and on this flight it is worth several GMCF points. 
R$_{POST}$ is likewise reported as a bound; no added markers survived
in view ($a = \bsAdjAddedUnits$), so Equation~\ref{eq:recall_post} is an upper bound.

\section{Discussion and Conclusion}
\label{sec:discussion}

This work demonstrates that \textit{BirdsEye} can transform expert observations made in the physical environment into image-space training annotations for aerial perception systems, while providing an explicit characterization of the geometric and sensor uncertainties governing that transformation. 
The resulting system combines direct RTK georeferencing, calibrated projective geometry, analytical uncertainty propagation, and field-deployed UAV imaging into a single annotation workflow. 
Across controlled flight experiments, the system achieved geospatial projection errors on the order of centimeters, with the largest errors occurring during sustained yawing maneuvers. 
These results establish that the proposed approach can provide sufficiently precise image-space correspondence for the demonstrated agricultural annotation task.

The uncertainty analysis provides an additional means of interpreting these empirical results. 
The first-order propagation model closely agrees with direct Monte Carlo simulation over the operational range of pose perturbations, while also identifying the parameters that most strongly influence pixel-space uncertainty. 
In particular, altitude uncertainty is a dominant contributor to our projection error, becoming increasingly important at lower operating altitudes. 
At sufficiently large perturbations, the analytical approximation diverges from the Monte Carlo result as higher-order effects become significant. 

The same propagation also runs backwards, which is what makes it a design tool
rather than only an analysis one. Because the trace of the projected covariance
is linear in the six per-axis pose variances, an annotation tolerance defines a
half-space of admissible noise budgets, evaluated at the worst-imaged point in
the frame. Screening a candidate sensor suite is therefore a single inner
product, the largest uniform inflation of a deployed suite has a closed form
(Equation~\ref{eq:kappa_star}), and a unique per-axis specification follows from
requiring each degree of freedom to contribute an equal share of the worst-case
pixel variance. That allocation is scale invariant, so the resulting tolerances
(Table~\ref{tab:design_bound_budget}) depend only on the imaging geometry and
the tolerance itself; they state a requirement on a platform rather than
describing the one we fly.

The price of that linearity is a condition that is sufficient but not tight, and
our own hardware sits close enough to the tolerance boundary to show where the
slack matters. At $h = 10$\,m the trace condition returns
$\kappa^\star = \bsKappaTraceTen$, rejecting the deployed budget by $8\%$, while
the exact worst-case eigenvalue condition accepts the same budget with
$\bsKappaInvTen\times$ headroom; at $h = 20$\,m
($\kappa^\star = \bsKappaTraceTwenty$) the two forms agree. The practical reading
is to allocate and screen with Equation~\ref{eq:design_bound_trace}, where
linearity in the design vector is the entire point, and to confirm a
near-boundary platform against $\sup_{\mathcal{X}} \lambda_{max}(\Sigma_u)$ at
the cost of an eigendecomposition per query. Table~\ref{tab:design_bound_budget}
is a design target and not a per-axis pass/fail test: conformance is decided by
Equation~\ref{eq:design_bound_trace} at the deployed budget, which this platform
satisfies, and an individual axis may exceed its balanced allocation while the
platform as a whole conforms.

A corresponding operating envelope can be established for the planar scene approximation. 
The current implementation represents the ground visible within each image as a locally planar surface at a depth estimated from radar altimetry. 
Our terrain-slope analysis indicates that terrain-induced reprojection error exceeds the motion-induced projection error at $\sim10^\circ$ of terrain slope, with error increasing rapidly beyond this regime. 
The flat-world approximation should therefore not be interpreted as an assumption that outdoor environments are globally planar. 
Rather, it is a local scene approximation whose validity depends on the relationship between terrain variation, camera altitude, field of view, and the allowable projection error. 
This characterization provides a practical criterion for determining when the planar model is adequate and when explicit three-dimensional scene geometry is warranted. 
The latter is a natural extension of the present architecture rather than a change to its fundamental annotation workflow.

The field deployment demonstrates the practical consequence of this coordinate-system inversion. 
Instead of requiring domain experts to identify targets repeatedly across individual aerial frames, experts can record observations directly in the environment and allow the projection pipeline to associate those observations with the corresponding imagery. 
The geoannotation hardware is assembled entirely from commercially available
components and free correction software, which matters for the intended use: a
workflow that requires bespoke instrumentation to reproduce is not one a grower or
a small research group can adopt.

In the reported timing experiments, two field workers generated a dataset containing 12,524 frames and 55,600 annotations across three field sites in roughly 12 hours. 
Relative to the manual image-annotation baseline used in this study, \textit{BirdsEye} increased per-worker annotation throughput by approximately 25.5$\times$. 
This comparison should be interpreted specifically as a comparison of the demonstrated end-to-end workflows rather than as a general claim about human annotation speed, since throughput depends on factors including image acquisition rate and the definition of the manual annotation task.

The downstream detection experiment tests the dataset, not the detector. 
A selection rule fixing \bsPreregArms{} operating points was sealed before any detection existed for the held-out site (Appendix~\ref{app:prereg}), and on that unseen flight the arms recover between \bsArmMaxPrecFlooredTestRecall{} and \bsArmMaxRecallTestRecall{} of the geotags in view under Equation~\ref{eq:inclusion}, against \bsArmLowestBurdenValRecall{} to
\bsArmMaxRecallValRecall{} on the validation flights. 
Recall therefore transferred, and it rose rather than fell ($\bsMeanDRecall$ on average across configurations). 
The GMCF moved the other way ($\bsMeanDPrec$), but that quantity is positive-unlabeled and its bias scales with survey density, and the validation flights are surveyed \bsSurveyDensityRatio$\times$ more densely than the test flight.
Therefore, it is not comparable across sites and no conclusion about detector quality follows from its decline. 
Human adjudication of the held-out detections raises GMCF on the max-F1 arm from \bsArmMaxFOneGeoPre{} to between \bsArmMaxFOnePostNeg{} and \bsArmMaxFOnePostPos, the range spanning the three tie-break conventions for clusters holding contradictory verdicts. 
This experiment supports the claim that the \textit{BirdsEye}-annotated dataset reached a size and complexity sufficient to carry modest models to good performance on an unseen site. 
It does not support any claim about the detector architecture, and none is made.

Several limitations remain. 
The present projection model relies on a locally planar representation of the scene and a single radar-derived depth estimate, making its accuracy increasingly dependent on terrain geometry as slopes and relief increase. 
The current uncertainty model is also based on first-order propagation and therefore does not capture higher-order effects under sufficiently large pose perturbations.
The design condition inherits that restriction directly, since it is built on
the same linearization, and it is stated at nadir for a locally planar scene;
oblique viewing geometry or a violated ground plane changes the gain vector and
therefore the admissible budget.
Finally, the field study evaluates the system primarily through a manual-annotation baseline and a single agricultural detection task. 
Broader comparisons across annotation strategies, sensor configurations, terrain types, and target classes would be necessary to establish performance across a wider range of field robotics applications.

These limitations point toward a direct extension of the present system: replacing the planar scene representation with explicit three-dimensional geometry and performing ray-surface intersection during annotation and detection projection. 
Such an extension preserves the central abstraction of \textit{BirdsEye} - human observations represented in world coordinates and automatically associated with sensor observations - while removing the principal geometric approximation identified in this study.

Overall, \textit{BirdsEye} demonstrates that geospatial annotation can serve as an effective intermediate representation between domain experts and aerial perception systems. 
By moving the annotation process from individual images into the physical environment, the system reduces the repeated image-space labeling burden while retaining explicit spatial correspondence and quantifiable geometric uncertainty. 
The combination of controlled geometric validation and multi-site field deployment provides evidence that this approach is practical for rapid construction of supervised datasets in outdoor robotic applications.


\authorcontributions{Conceptualization, M. Masters. and N. Bender.; methodology, M. Masters, A. Korycki, N. Bender, T.L. Altaffer., and N. Kuipers; software, M. Masters.; validation, M. Masters. A. Korycki, and T.L. Altaffer.; formal analysis, M. Masters.; investigation, M. Masters, T.L. Altaffer, and N. Kuipers.; resources, S. McGuire and C. Josephson.; data curation, M. Masters.; writing---original draft preparation, M. Masters and A. Korycki.; writing---review and editing, M. Masters., S. McGuire, C. Josephson, and A. Korycki; visualization, M. Masters and A. Korycki.; supervision, S. McGuire. and C. Josephson; project administration, C. Josephson and S. McGuire.; funding acquisition, S. McGuire, C Josephson, and M. Masters. All authors have read and agreed to the published version of the manuscript.}

\funding{This research was funded by the Engineering for Precision Water and Crop Management Program, project award no. 2023-67022-40557, from the U.S. Department of Agriculture’s National Institute of Food and Agriculture and by UC Santa Cruz Agricultural Experiment Station funding provided by the state of California.}

\dataavailability{The code supporting the conclusions of this article is openly available at \url{https://github.com/harelab-ucsc/birdseye/tree/pose-bound}. The raw data supporting the conclusions of this article will be made available by the authors on request.} 

\acknowledgments{We would like to thank Jake Lee, Derick Mathews, and Andrea Arreortua for their piloting support and design work, as well as Pie Ranch and Jacobs Farms / del Cabo, for volunteering their properties as test sites for \textit{BirdsEye}.  During the preparation of this manuscript, the authors used GPT-5 and GPT-6 for the purposes of manuscript revision and software review. Claude Opus 5 generated the GUI used in the human adjudication work, as well as the experimental harness for the applied analytical bound. The authors have reviewed and edited the output and take full responsibility for the content of this publication.}

\conflictsofinterest{The authors declare no conflicts of interest.} 

\appendixtitles{yes} 
\appendixstart
\appendix

\section[\appendixname~\thesection]{Propagation of Pose Uncertainty into Pixel Space}
\label{sec:methods_sensitivity}
We present a first-principles analysis of pose uncertainty propagation through our pipeline. 
This allows us to provide bounds for the operational envelope of the system and, within this envelope, to provide a fast estimate of system calibration quality.
To do so, we first consider our projective function:
\begin{equation}
\label{eqn:projection}
    \underline{u} = \pi(K T_{wc} \underline{X}_w) = \pi(\hat{\underline{u}})
\end{equation}
where $\underline{u}$ represents the projected coordinates of a 3D point, $\pi(.)$ represents perspective division, $K$ represents the 3x4 camera intrinsics matrix, $T_{wc}$ represents the rigid-body transform from the world frame to the camera frame, $\underline{X}_w$ and $\underline{X}_c$ represent a point in the world frame and the camera frame, respectively, and $\hat{\underline{u}}$ represents an un-normalized projected point.

To inject noise into Eq.~\ref{eqn:projection}, we recognize that we require the exponential map between the $\mathfrak{se}(3)$ Lie algebra, representing poses as 6-D vectors, and the $SE(3)$ Lie group, representing the poses in 3D space.
If we consider left-perturbations of the pose matrix, where our perturbation is $\underline{\xi} \sim \mathcal{N}(0, \Sigma_0) \in \mathbb{R}^6$, we have:
\begin{equation}
\label{eqn:perturbation}
    T_{wc} = \Delta T \times T_0 = \exp(\underline{\xi} \verb!^!) \times T_0
\end{equation}
Here, we take $\Sigma_0$ to be a diagonal matrix whose entries are provided by system sensor datasheets, $\Delta T$ is our perturbation, and $T_0$ is an arbitrary nominal pose.
Additionally, if we further specify that $ \underline{\xi}^T = [\rho^T \;\; \phi^T] $, where $\rho$ is a 3D translation vector and $\phi$ is a 3D rotation vector, then the hat operator (`` $\underline{\xi}\verb!^!$ ") is defined as follows for the 6-DoF pose vector:
\begin{equation}
    \underline{\xi}\verb!^! = \begin{bmatrix}
        \phi\verb!^! & \rho \\ 
        \underline{0}^T & 0
    \end{bmatrix} \in \mathbb{R}^{4\times4}
\end{equation}
The ``hat" operator used on $\phi$ within the definition of $\underline{\xi}\verb!^!$ is defined as 
\begin{equation}
    \label{eq:phihat}
    \phi\verb!^! = 
    \begin{bmatrix} 0 & -\phi_3 & \phi_2 \\ \phi_3 & 0 & -\phi_1 \\ -\phi_2 & \phi_1 & 0 \end{bmatrix} 
    \in \mathbb{R}^{3\times3}
\end{equation}
To understand the overloading of this operator, see~\cite{Gao2021}.
This noise injection scheme assumes that rotation uncertainty is small, such that the tails of the normal distribution produce negligible aliasing due to the periodicity of angular quantities.

Now, Eq.~\ref{eqn:projection} becomes:
\begin{equation}
\label{eqn:proj_pert}
    \underline{u} = \pi(K \exp(\underline{\xi} \verb!^!) T_0 \underline{X}_w) = \pi(\hat{\underline{u}}(\underline{\xi}))
\end{equation}

Our analysis focuses on projective error $\underline{\varepsilon} = \underline{u} - \underline{u}_{True}$, where $\underline{u}$ is our model's prediction and $\underline{u}_{True}$ is the ground truth.
Taking the Taylor expansion of $\underline{u}$ and discarding all terms of second order and above, we have:
\begin{equation}
\label{eqn:error_base}
    \underline{\varepsilon} = \left[\underline{u}(0) + \frac{\partial \underline{u}}{\partial \underline{\xi}}(\underline{\xi} - 0) + ...\right] - \underline{u}_{True}
\end{equation}
Noting that $\underline{u}(0) = \underline{u}_{True}$, this leaves:
\begin{equation}
    \underline{\varepsilon} = \frac{\partial \underline{u}}{\partial \underline{\xi}}(\underline{\xi} - 0) = J_\xi \underline{\xi}
\end{equation}
where $J_\xi$ is the Jacobian of $\underline{u}$ with respect to $\underline{\xi}$.
Therefore, $\underline{\varepsilon} \sim \mathcal{N}(0, \Sigma_u = J_\xi\Sigma_0J_\xi^T)$ under these assumptions.

We now require $J_\xi$ to attain a closed-form expression for the covariance matrix of the projection error. 
By the chain rule:
\begin{equation}
    J_\xi = \frac{\partial \underline{u}}{\partial \underline{\hat{u}}} \frac{\partial \underline{\hat{u}}}{\partial \underline{\xi}} = \frac{\partial \underline{u}}{\partial \underline{\hat{u}}} K \frac{\partial \underline{X}_c}{\partial \underline{\xi}}
\end{equation}

Since $\underline{u} = \left[ \hat{u}_1/\hat{u}_3 \quad \hat{u}_2/\hat{u}_3 \quad 1 \right]^T$, we have:
\begin{equation}
    \frac{\partial \underline{u}}{\partial \underline{\hat{u}}} = \begin{bmatrix}
        1/\hat{u}_3 & 0 & -\hat{u}_1/\hat{u}_3^2 \\
        0 & 1/\hat{u}_3 & -\hat{u}_2/\hat{u}_3^2 \\
        0 & 0 & 0
    \end{bmatrix}
\end{equation}

To get an expression for $\frac{\partial \underline{X}_c}{\partial \underline{\xi}}$, we follow the derivation of~\cite{Gao2021}, which gives:
\begin{equation}
    \label{eq:dxc_dxi}
    \frac{\partial \underline{X}_c}{\partial \underline{\xi}} = \begin{bmatrix}
        \mathrm{I_{3x3}} & (R \underline{X}_w + \underline{t})\verb!^! \\
        \underline{0}^T & \underline{0}^T
    \end{bmatrix}
\end{equation}
In Equation~\ref{eq:dxc_dxi}, the hat operator uses the definition of Equation~\ref{eq:phihat}. The authors use a first-order approximation of the exponential map to arrive at this result, and $R$ and $\underline{t}$ represent the rotation and translation associated with $T_0$. 
Thus:
\begin{equation}
    \label{eq:jacobian_xi}
    J_\xi = \begin{bmatrix}
        1/\hat{u}_3 & 0 & -\hat{u}_1/\hat{u}_3^2 \\
        0 & 1/\hat{u}_3 & -\hat{u}_2/\hat{u}_3^2 \\
        0 & 0 & 0
    \end{bmatrix} K 
    \begin{bmatrix}
        \mathrm{I_{3x3}} & (R \underline{X}_w + \underline{t})\verb!^! \\
        \underline{0}^T & \underline{0}^T
    \end{bmatrix}
\end{equation}

Section~\ref{sec:bound} applies this Jacobian to derive the design bounds, and
Section~\ref{sec:results_sensitivity} validates the linearization against Monte
Carlo simulation.

\section[\appendixname~\thesection]{Detection Metrics and Their Estimands}
\label{app:metrics}

In our case study, detections are evaluated in the ground plane rather than in the image, because a burrow seen from ten frames is one burrow and not ten. 
Every detection in every processed frame is back-projected through its own frame's pose, the resulting points are deduplicated onto a \bsDedupCm~cm grid, and the survivors are clustered with DBSCAN at radius $\varepsilon$ and minimum density $m$. 
A cluster therefore spans frames by construction, and the unit of precision is the cluster, not the detection. 
Cluster identity depends on $(\varepsilon, m)$; values measured at different clustering parameters are
not comparable, and Section~\ref{sec:holdout} reports them separately
for that reason.

Recall is measured against the RTK geotag survey. 
A geotag counts as recovered if it falls within $\varepsilon_{\mathrm{pred}}$ of a core point of some cluster, and the denominator is the set of geotags that were \emph{in view}: those satisfying the in-frame inclusion test of Equation~\ref{eq:inclusion} in at least one processed frame. 
Under Equation~\ref{eq:inclusion}, \bsTestInView{} of the \bsTestGeotags{} geotags on the held-out flight are in view.

We define a precision-like quantity, referred to here as geotag-matched cluster fraction (\textbf{\bsGmcfName}): the fraction of clusters containing at least one in-view geotag. 
If adjudication has occurred, we expand this definition to clusters carrying a \texttt{confirmed-positive} verdict.

\begin{equation}
    \label{eq:gmcf}
    GMCF_{\mathrm{PRE}} \;=\; \frac{n_{hit}}{n_{clusters}} \quad GMCF_{\textrm{POST}} \;=\; \frac{n_{hit} + c}{n_{clusters}}
\end{equation}
where $n_{hit}$ is the number of clusters containing at least one geotag, $n_{clusters}$ the number of detection clusters produced at this operating point, and $c$ is the number of human-confirmed positives.

\textbf{\bsGmcfName{} is not precision.}
The geotag survey does not exhaustively enumerate burrows, so a cluster sitting on a real but unsurveyed burrow is counted against it, and because a more sensitive operating point produces more clusters over the same fixed survey, the penalty grows with detection rate. 
\bsGmcfName{} is a lower bound on precision whose bias scales with survey density; it is valid for relative comparison within one site and invalid across sites. 
This is a positive-unlabeled evaluation problem in the sense of Bekker and Davis~\cite{bekker2020pu}, and it is the reason the held-out comparison in Section~\ref{sec:holdout} treats recall and \bsGmcfName{} asymmetrically.

Two statistical arms are reported throughout Section~\ref{sec:holdout}. 
The \textbf{PRE} arm is geometry only: a cluster is a true positive if a geotag reached it. 
Its recall is an unbiased point estimate under the assumption that survey omission is independent of detectability.
That assumption is defensible here on physical grounds.
The geotags come from a walking RTK survey on the ground, the detector reads aerial imagery, and the two miss mechanisms are unrelated: what makes a burrow easy to walk past is not what makes it hard to see from altitude. 

The \textbf{POST} arm admits the human adjudication and redefines a true positive as
\emph{human-confirmed positive or geometrically matched}. 
Enlarging the positive
class enlarges the recall denominator with it, so the two arms cannot share one
recall:
\begin{equation}
    \label{eq:recall_post}
    R_{\mathrm{PRE}} \;=\; \frac{n_{rec}}{n_{den}}, \quad R_{\mathrm{POST}} \;=\; \frac{n_{\mathrm{rec}} + c}{n_{\mathrm{den}} + c + a}
\end{equation}
where $n_{rec}$ is the number of geotags within $\varepsilon_{pred}$ of a detection cluster, $n_{den}$ is the total number of geotags which the camera saw (by Equation~\ref{eq:inclusion}), $c$ is defined as in Equation~\ref{eq:gmcf}, $a$ is the number of distinct targets a reviewer marked where the detector never fired and the surveyor never geotagged. 
With $a = 0$ this is a strict \emph{upper} bound and not a point estimate, because $c$ is censored: unlabeled positives are discoverable only where the detector fired, so
adjudication can only push the ratio up. 
No added markers were placed in this review pass, so every R$_{POST}$ reported here is such a bound and is printed with ``$\geq$" as a reminder.

No F1-score (the harmonic mean of a detector's precision and recall scores) is reported in the geospatial arm. 
Where an F1 appears in this paper it is the pooled validation quantity defined over the screening flights, it was used for checkpoint selection only, and it is defined in pixel space.
Precision in our geospatial domain is defined over clusters while recall is over geotags.
Since the units of precision and recall are not identical, and since one cluster can absorb several geotags, the two ratios have different denominator units and meanings.
Therefore, their harmonic mean is a category error. 

A single cluster can contain both a \texttt{confirmed\_real} and a \texttt{hard\_negative} judgment, because human review is not perfect, and two views of the same detection target may lead a human to different conclusions.
A secondary source of such labeling conflict is that DBSCAN chains at $\varepsilon = \bsArmMaxFOneEps$~m and two physically distinct spots a few decimeters apart can merge under reviewer propagation of spatial cluster labels.
Which verdict then governs the cluster is fixed by a tie-break convention, not by data, and no further review resolves it. 
Our conventions are \emph{positive-wins}, \emph{majority-wins}, and \emph{hard-negative-wins} and the spread across them is part of the result rather than an error bar on it. 

\section[\appendixname~\thesection]{Model Selection and Held-Out Evaluation}
\label{ap:model_selection}

\subsection[\appendixname~\thesection]{Checkpoint Screening}
\label{app:screen}

\begin{figure}[ht]
  \centering
  \includegraphics{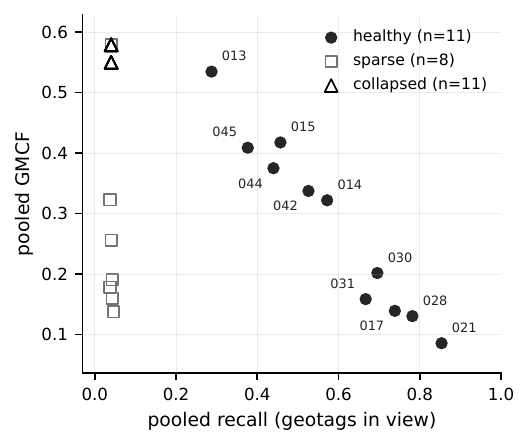}
  \caption{Pooled recall against pooled \bsGmcfName{} for all
  \bsScreenCkpts{} screened checkpoints, micro-averaged over the
  \bsScreenFlights{} validation flights. Filled circles are the
  \bsScreenHealthy{} checkpoints that passed both health guards and are tagged
  with their checkpoint identifier; open squares are sparse and open triangles
  collapsed.}
  \label{fig:frontier}
\end{figure}

\begin{figure}[H]
  \centering
  \includegraphics[width=\linewidth]{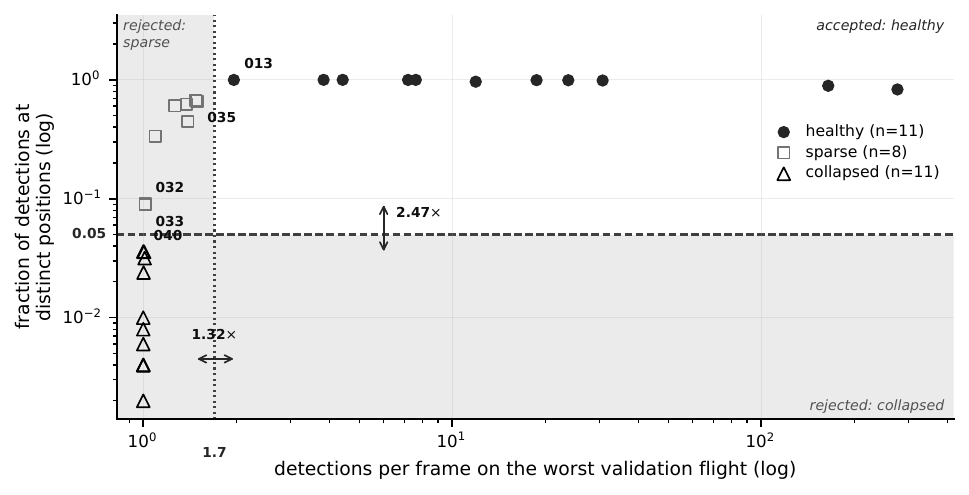}
  \caption{The two health guards as a decision rule, for all
  \bsScreenCkpts{} screened checkpoints. Both axes are logarithmic. Shaded
  strips are the rejected regions: below the distinct-position cut
  (\bsCollapseCut, dashed) a checkpoint is collapsed, and left of the
  detection-rate cut (\bsSparseCut{} det/frame on its worst validation flight,
  dotted) it is too sparse to score. 
  The double-headed arrows are the margins -- the ratio between
  the nearest checkpoint on either side of each cut -- and only the
  cut-bounding checkpoints are labeled, because they are the ones whose values
  decide where a cut may sit.
  A cut placed on a \bsScreenCkpts-point sample is an argument about its neighbors, not a natural boundary.
  }
  \label{fig:guards}
\end{figure}

All \bsScreenCkpts{} training checkpoints were scored on
\bsScreenFlights{} validation flights at a fixed confidence threshold of
\bsScreenThresh, over \bsScreenBlockFrames{}-frame blocks
(\bsScreenFrames{} frames in total, \bsScreenInView{} geotags in view under
Equation~\ref{eq:inclusion}). Results are micro-averaged: numerators and denominators are
summed across flights and divided once. The flights differ several-fold in how
many geotags they put in view, so averaging per-flight ratios would weight a
sparse flight equally with a dense one and report a mean of incommensurable
quantities. 
The results of these sweeps are presented in Figure~\ref{fig:frontier}
.
Two health guards are applied before any checkpoint is ranked, on two
independent axes of failure (Figure~\ref{fig:guards}). The first is a
\emph{distinct-position} test: the fraction of a checkpoint's detections that
land at distinct image positions, cut at \bsCollapseCut. It catches a
checkpoint that has collapsed to a near-constant response, emitting the same
pixel in frame after frame. The second is a \emph{detection-rate} test:
detections per frame on the checkpoint's \emph{worst} validation flight, cut at
\bsSparseCut. It catches a checkpoint that fires too rarely for any ratio
computed from it to mean anything. Neither subsumes the other, and the two
groups in Figure~\ref{fig:guards} fail exactly one test each: a checkpoint can
fire constantly at one pixel, and it can fire at well-spread positions but
almost never. Of \bsScreenCkpts{} checkpoints, \bsScreenHealthy{} passed both,
\bsScreenCollapsed{} were collapsed and \bsScreenSparse{} were too sparse to
score, so \bsScreenExcluded{} were excluded before any operating point was
considered. A third filter, on cluster geometry rather than on detector output,
is applied later to configurations rather than checkpoints
(Appendix~\ref{app:sweep}).

\begin{table}
  \caption{The \bsScreenHealthy{} healthy checkpoints, micro-averaged over the
  \bsScreenFlights{} validation flights, sorted by pooled detections per frame
  descending. Review load is relative to the lightest checkpoint. The field
  spans \bsBurdenSpan$\times$ in review load for \bsRecallSpan$\times$ in
  recall, which is an operational trade and not a ranking: no row in this table
  is a recommendation. F1 is the pooled validation quantity used for selection
  only (Appendix~\ref{app:metrics}). The remaining
  \bsScreenExcluded{} of \bsScreenCkpts{} checkpoints were excluded by the
  health guards (\bsScreenCollapsed{} collapsed, \bsScreenSparse{} sparse).}
  \label{tab:screen}
\begin{tabularx}{\textwidth}{lCCCCC}
\toprule
Ckpt & det/frame (pooled) & Recall & \bsGmcfName{} & F1 (selection only) & Relative review load \\
\midrule
028 & 249.67 & 0.782 & 0.130 & 0.224 & 66.9$\times$ \\
021 & 176.80 & 0.853 & 0.086 & 0.156 & 47.4$\times$ \\
017 & 43.21 & 0.739 & 0.139 & 0.234 & 11.6$\times$ \\
031 & 34.46 & 0.667 & 0.159 & 0.256 & 9.2$\times$ \\
030 & 30.03 & 0.695 & 0.202 & 0.313 & 8.0$\times$ \\
044 & 14.20 & 0.440 & 0.375 & 0.405 & 3.8$\times$ \\
014 & 14.07 & 0.572 & 0.322 & 0.412 & 3.8$\times$ \\
042 & 13.70 & 0.526 & 0.337 & 0.411 & 3.7$\times$ \\
015 & 7.17 & 0.457 & 0.418 & 0.436 & 1.9$\times$ \\
045 & 6.38 & 0.376 & 0.409 & 0.392 & 1.7$\times$ \\
013 & 3.73 & 0.287 & 0.535 & 0.374 & 1.0$\times$ \\
\bottomrule
\end{tabularx}

\end{table}

\subsection[\appendixname~\thesection]{Operating-Point Sweep}
\label{app:sweep}

Clustering parameters were swept over four axes -- confidence threshold,
$\varepsilon$, $m$, and $\varepsilon_{\mathrm{pred}}$ --
giving \bsSweepConfigs{} pooled configurations. Of these, \bsSweepCredible{}
satisfy the geometric credibility filter (mean geotags per hit cluster
$\leq \bsCredMaxGeotagsPerCluster$ and mean median cluster extent
$\leq \bsCredMaxMedianExtent$~m); \bsSweepNoHitCluster{} produced no hit
cluster at all. The filter is not cosmetic. Without it the Pareto front is
occupied by chained configurations, which post high \bsGmcfName{} \emph{and}
high F1 precisely because merging shrinks the cluster denominator as fast as it
inflates the matched numerator, while the clusters themselves absorb tens of
geotags across tens of meters and are useless for locating a burrow.

\subsection{Pre-registration}
\label{app:prereg}

A selection rule was written and cryptographically sealed before any detection existed for the held-out flight. 
The rule pins \bsPreregArms{} confirmatory arms (Table~\ref{tab:arms}), each fixing both a checkpoint and its complete operating point, and registers \bsPreregConfigs{} configurations in total across \bsPreregCkpts{} checkpoints.
The configurations which are not pinned as confirmatory are pre-registered as descriptive. 

\begin{adjustwidth}{-\extralength}{0cm}

\reftitle{References}

\bibliography{refs.bib}

\PublishersNote{}
\end{adjustwidth}
\end{document}